\documentclass[conference]{IEEEtran}
\IEEEoverridecommandlockouts
\usepackage{cite}
\usepackage{amsmath,amssymb,amsfonts}
\usepackage{bm}
\usepackage{algorithm}
\usepackage{algorithmic}
\usepackage{graphicx}
\usepackage{textcomp}
\usepackage{xurl}
\usepackage{microtype}
\usepackage[font=small,skip=3pt]{caption}
\usepackage{booktabs}
\usepackage{multirow}
\usepackage{amsthm}
\usepackage{etoolbox}
\theoremstyle{definition}
\newtheorem{proposition}{Proposition}
\newtheorem{corollary}{Corollary}
\theoremstyle{remark}

\usepackage[capitalise]{cleveref}
\crefname{section}{Section}{Sections}\crefname{subsection}{Section}{Sections}
\crefname{subsubsection}{Section}{Sections}\crefname{table}{Table}{Tables}
\crefname{figure}{Fig.}{Figs.}\crefname{proposition}{Proposition}{Propositions}
\crefname{corollary}{Corollary}{Corollaries}\crefname{appendix}{Appendix}{Appendices}
\crefformat{equation}{(#2#1#3)}
\crefrangeformat{equation}{(#3#1#4)--(#5#2#6)}
\crefmultiformat{equation}{(#2#1#3)}{ and~(#2#1#3)}{, (#2#1#3)}{ and~(#2#1#3)}
\newcommand{\PARstart}[2]{#1#2}
\newcommand{\fitwidth}[1]{\resizebox{\ifdim\width>\linewidth \linewidth\else\width\fi}{!}{#1}}
\newcommand{\CodeRepo}{\url{https://github.com/phamvunhatminh/AV-DFER-OT}}

\newcommand{\MmaDfewBase}{65.89 / 76.87}
\newcommand{\MmaDfewScr}{65.80 / 76.65}
\newcommand{\MmaDfewPool}{65.94 / 76.63}
\newcommand{\MmaMafwBase}{43.48 / 56.82}
\newcommand{\MmaMafwScr}{43.59 / 56.74}
\newcommand{\MmaMafwPool}{43.28 / 56.58}
\newcommand{\AbvDfewBase}{66.68 / 77.82}
\newcommand{\AbvDfewScr}{67.56 / 78.47}
\newcommand{\AbvDfewPool}{67.39 / 78.22}
\newcommand{\AbvMafwBase}{43.80 / 57.90}
\newcommand{\AbvMafwScr}{44.41 / 58.60}
\newcommand{\AbvMafwPool}{44.38 / 58.66}

\newcommand{\gScrDfew}{0.87}   %
\newcommand{\gScrMafw}{0.61}   %
\newcommand{\gPoolDfew}{0.70}  %
\newcommand{\gPoolMafw}{0.58}  %

\newcommand{\mScrDfew}{0.17}   %

\newcommand{\MmaBound}{0.12}

\newcommand{\SlotRatioMin}{0.985}   %
\newcommand{\FramePartMax}{1.8\%}   %
\newcommand{\AbvSlotLo}{0.91}  %
\newcommand{\AbvSlotHi}{0.97}  %

\newcommand{\FrameSpreadMax}{2.2\%}   %
\newcommand{\SlotSpreadMin}{24\%}     %

\newcommand{\ColKappaMax}{0.21}       %
\newcommand{\RowTopSlotMin}{0.54}     %
\newcommand{\RowFramePartMax}{0.5\%}  %

\newcommand{\RowResidMax}{8\times10^{-3}}        %
\newcommand{\ColResidMax}{1.3\times10^{-6}}      %
\newcommand{\PriorRowResidMax}{8.3\times10^{-2}} %

\newcommand{\PriorDfewFrameUpdate}{8.0\%}   %
\newcommand{\PriorCremaFrameUpdate}{9.6\%}  %
\newcommand{\PriorScrDfew}{-0.20}           %
\newcommand{\PriorScrCrema}{+0.19}          %
\newcommand{\PriorPoolCrema}{+0.00}         %

\newcommand{\ConfirmCIHalfLo}{0.3}
\newcommand{\ConfirmCIHalfHi}{1.3}

\newcommand{\PadWindow}{5.12~s}
\newcommand{\PadShareDfew}{40.3\%}
\newcommand{\PadShareMafw}{34.6\%}
\newcommand{\MisplacedClips}{1{,}544}

\newcommand{\FoldFactor}{1.5}
\newcommand{\MDEScrPool}{1.0}
\newcommand{\ScrPoolPower}{0.80}
\newcommand{\ScrPoolUpperMax}{1.2}
\newcommand{\FoldResolved}{eleven}
\newcommand{\FoldTotal}{86}
\newcommand{\FoldExpected}{nine}
\newcommand{\LooGainKept}{five}
\newcommand{\LooContrastLost}{three}
\newcommand{\LooSwapLost}{four}

\newcommand{\MixDrop}{3.8}
\newcommand{\TempDepthSpread}{1.04}
\newcommand{\AxesScrLate}{0.75}
\newcommand{\AxesScrLateCI}{[-1.25, -0.25]}
\newcommand{\FoldKept}{three}

\newcommand{\ScrFpDelta}{0.50}            %
\newcommand{\ScrFpCI}{[-0.02, +1.02]}     %
\newcommand{\PrecShiftDelta}{0.47}        %
\newcommand{\PrecShiftCI}{[+0.21, +0.73]} %

\newcommand{\ReproGapLo}{0.66}
\newcommand{\ReproGapHi}{0.84}

\newcommand{\epsdef}{0.05}

\newcommand{\PermAbvDfew}{-0.20}
\newcommand{\PermAbvDfewCI}{[-0.49, +0.09]}
\newcommand{\PermMinusSwap}{+0.51}
\newcommand{\PermMinusSwapCI}{[+0.15, +0.88]}
\newcommand{\PermMinusSwapMag}{0.51}
\newcommand{\PermMinusSwapWideCI}{[-0.03, +1.06]}
\newcommand{\PermOtherMax}{0.09}

\newcommand{\PadMaskDrop}{0.25}
\newcommand{\AxesCIHalfLo}{0.5}
\newcommand{\AxesCIHalfHi}{1.4}

\newcommand{\EpochCount}{125}
\newcommand{\EpochBase}{191.8}
\newcommand{\EpochScr}{209.7}
\newcommand{\EpochPool}{195.8}
\newcommand{\EpochScrPct}{9.3}
\newcommand{\EpochPoolPct}{2.1}

\newcommand{\PerClassCount}{seven}
\newcommand{\PerClassResolved}{one}
\newcommand{\PerClassDisgust}{2.76}
\newcommand{\PerClassDisgustCI}{[+0.01, +5.51]}
\newcommand{\PerClassDisgustClips}{145}
\newcommand{\PerClassDown}{two}
\newcommand{\SwapAbvDfewAllCI}{[-1.31, -0.12]}   %
\newcommand{\SwapAbvDfewAllMag}{0.72}   %
\newcommand{\gScrDfewCI}{[+0.43, +1.32]}   %
\newcommand{\gScrMafwCI}{[-0.27, +1.49]}   %
\newcommand{\mScrDfewCI}{[-0.02, +0.36]}   %
\newcommand{\mScrMafwCI}{[-0.46, +0.52]}   %
\newcommand{\PlanColPoolDfew}{0.96}   %
\newcommand{\PlanColPoolMafw}{0.58}   %
\newcommand{\DevAbvBaseScr}{1.34}   %
\newcommand{\DevMmaPoolBase}{0.82}   %
\newcommand{\PlanScrPoolBound}{0.54}   %
\newcommand{\SwapMmaAllBound}{0.07}   %
\newcommand{\ProdAbvDfewAllCI}{[-1.27, -0.13]}   %
\newcommand{\ProdAbvDfewAllMag}{0.70}   %
\newcommand{\ProdAbvDfewAllWideCI}{[-1.55, +0.15]}   %
\newcommand{\ProdResolvedCount}{one}   %
\newcommand{\ProdSettingCount}{five}   %
\newcommand{\ProdOtherBound}{0.16}   %
\newcommand{\HLevel}{0.10}
\newcommand{\HMedianBound}{0.059}   %
\newcommand{\HShareMin}{64.7\%}   %
\newcommand{\HPairs}{590{,}112}   %
\newcommand{\HFirstBound}{0.001}   %
\newcommand{\HMmaEarlyBound}{0.06}   %
\newcommand{\HFinalMmaLo}{0.21}   %
\newcommand{\HFinalMmaHi}{0.30}   %
\newcommand{\HFinalAbvLo}{0.18}   %
\newcommand{\HFinalAbvHi}{0.19}   %
\newcommand{\SwapCILevel}{90\%}
\newcommand{\SwapFoldCount}{5}

\begin{document}
\title{\LARGE Separating Module Gain From Transport Plan Contribution\\in Audio-Visual Dynamic Facial Expression Recognition}
\author{\IEEEauthorblockN{Vu Nhat Minh Pham, Cong Khanh Nguyen, Minh Tuan Le, Van Thong Huynh}
\IEEEauthorblockA{Department of AI, FPT University, Ho Chi Minh Campus, Vietnam\\
\texttt{\{minhpvnse180448, khanhncse190752, tuanlmse184475\}@fpt.edu.vn}, \texttt{thonghv4@fe.edu.vn}}}
\maketitle
\begin{abstract}
A cross-modal routing module can improve over its backbone while its transport plan
accounts for no demonstrable part of that improvement. In ABVnet with DFEW, Sinkhorn
cross-modal routing (SCR) exceeds the retrained baseline by \gScrDfew{} unweighted average
recall (UAR) points, a resolved difference, yet sits \mScrDfew{} points above an individually
trained, parameter-matched pooled variant, an unresolved difference that establishes no
equivalence; under the development protocol the comparison with the baseline reverses.
Separating the two needs comparisons a module-against-backbone test cannot supply:
parameter-matched variants changing only the plan rule, and interventions replacing the
plan without retraining. The evaluation
covers MMA-DFER and ABVnet on DFEW and MAFW, and MMA-DFER on CREMA-D, with a single training
run per configuration and paired fold intervals. ABVnet with DFEW
motivated the follow-up after the broader results were available. Replacing the plan everywhere with a
uniform one reduces UAR by \SwapAbvDfewAllMag{}, whereas permuting its rows gives an unresolved difference of \PermAbvDfew{} UAR points; permutation is less damaging by
\PermMinusSwapMag{} UAR points, a resolved contrast. Replacing the plan with the product of its own
marginals, preserving both marginals and removing only frame-specific mixtures, reduces
UAR by \ProdAbvDfewAllMag{}. These reductions concern reliance at inference; attribution of the
module gain to the transport plan remains unestablished. Under the \FoldFactor{}
fold-dependence widening, the baseline gain remains resolved, whereas the replacement
reductions and the paired contrast are unresolved.
\end{abstract}

\section{Introduction}
\label{sec:intro}

\begin{figure}[t]
\centering
\includegraphics{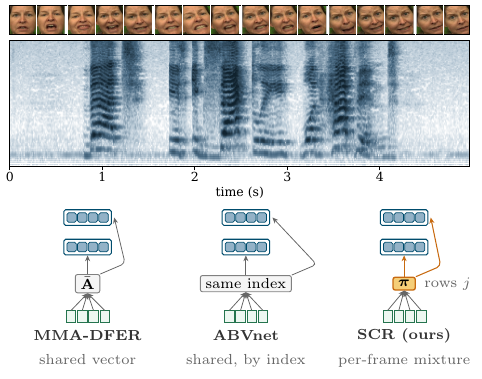}
\caption{The audio each frame receives. Above: the sampled frames at their sampling times over the
log-mel filterbank of the same CREMA-D clip. Below: what each design permits, not what it
learns.}
\label{fig:concept}
\end{figure}

\PARstart{D}{ynamic} facial expression recognition (DFER) in the wild is increasingly treated as an audio-visual task. A common design freezes pretrained encoders and trains only the modules that pass information between the streams~\cite{chumachenko2024mmadfer,yang2026abvnet,ning2026parameter,zhang2026bhgap}. The rule by which those modules exchange information decides which audio reaches each frame. MMA-DFER~\cite{chumachenko2024mmadfer} broadcasts the mean audio latent to every token of every frame, whereas ABVnet~\cite{yang2026abvnet} pairs adapted audio and image tokens by index inside every frame. Neither native exchange pairs frames and audio by time (\Cref{fig:concept}).

Comparing a complete transport-based alignment module with its backbone evaluates the routing path and the transport plan together~\cite{han2022mmalign,li2025alignmamba,ma2023attsinkhorn}. An improvement in that comparison therefore establishes a benefit from the complete addition under the evaluated conditions, but does not isolate an advantage of the transport plan.

Isolating an advantage of the transport plan calls for comparisons within the routing module. Such comparisons follow the rationale for retaining the surrounding architecture when comparing alternative token mixers~\cite{yu2022metaformer,leethorp2022fnet}, while addressing a narrower question than whole-model interaction analyses~\cite{hessel2020emap,liang2023pid}; \Cref{sec:related} places this study against both literatures in full. This study addresses the within-module attribution question using an individually trained, parameter-matched pooled variant and paired interventions on computed transport plans.

Individually trained comparisons and paired interventions answer different attribution questions. Training a variant whose only difference is the rule forming the plan shows what the plan contributes to a model built around that rule, whereas replacing the plan without retraining shows what a trained model has come to rely on. Neither kind of difference can be added or subtracted to decompose the gain over the baseline. The contrast between row permutation and uniform replacement tests whether they differ, without treating a comparison whose interval includes zero as evidence either for a benefit of the transport plan or against one.

In this paper, we conduct a controlled empirical study using Sinkhorn cross-modal routing (SCR) as the instrument, a routing module that forms a transport plan between frame descriptors and audio time slots. One shared module serves both kinds of comparison, which separates evidence about the transport plan from evidence about the complete module (\Cref{sec:method}). The evaluation covers MMA-DFER and ABVnet on DFEW~\cite{jiang2020dfew} and MAFW~\cite{liu2022mafw}, with an additional MMA-DFER evaluation on CREMA-D~\cite{cao2014crema}.
\begin{itemize}
\item We show that SCR improves UAR over the retrained baseline in ABVnet with DFEW by \gScrDfew{} points, while the additional benefit over the individually trained, parameter-matched pooled variant remains unresolved at \mScrDfew{} points; attribution of the module gain to the transport plan is therefore not established (\Cref{tab:comparisons}).
\item We show, by paired interventions without retraining in the same trained models, that row permutation is less damaging than uniform replacement by \PermMinusSwapMag{} UAR points, and that replacing the plan with the product of its own marginals reduces UAR by \ProdAbvDfewAllMag{} points while leaving both measured marginals in place; neither difference survives the \FoldFactor{} fold-dependence widening, and the benefit of the original frame--slot correspondence remains unestablished (\Cref{tab:comparisons}).
\end{itemize}

\Cref{sec:method} defines the comparison design and the shared routing module, and \Cref{sec:expr} brings together the evaluation protocol, the five comparisons, and the supporting mechanism measurements. \Cref{sec:discussion} states the scope and limitations of the attribution argument and closes with the conclusion in \Cref{sec:conclusion}. The routing implementation, the backbone reproduction and the design-axis comparisons are given in the appendix.
\section{Related Work}
\label{sec:related}

\subsection{Dynamic Facial Expression Recognition}

Vision-only DFER treats expression recognition as a problem of representing facial dynamics, on the shared assumption that the visual stream suffices for clip-level classification. Self-supervised video pretraining~\cite{sun2023maedfer,sun2025svfap}, vision-language transfer~\cite{zhao2023dferclip}, and adaptation of static expression models~\cite{chen2025s2d,chen2026s4d} differ in where those representations come from rather than in that assumption. DFEW~\cite{jiang2020dfew} and MAFW~\cite{liu2022mafw} are the standard in-the-wild benchmarks, both with five-fold protocols over film and television clips. The strength of these methods is what makes the added value of audio worth isolating.

Audio-visual DFER treats the task as combining modality representations, on the premise that both streams carry evidence. Reliability-based fusion weights the streams by their estimated reliability~\cite{hassanie2026ramer}, and joint masked-autoencoder pretraining learns both modalities together~\cite{sun2024hicmae,wu2025avfmae}. A third route concentrates training in small modules between frozen pretrained encoders. MMA-DFER~\cite{chumachenko2024mmadfer} and ABVnet~\cite{yang2026abvnet} pair MAE-Face~\cite{ma2023maeface} with AudioMAE~\cite{huang2022audiomae}, as do MFAdapter~\cite{ning2026parameter} and BHGap~\cite{zhang2026bhgap}; CLAIP-Emo~\cite{chen2026claipemo} instead adapts frozen language-supervised encoders with low-rank updates. These routes share a reliance on pretrained representations and differ in which components train. A comparison that changes the encoders or the fusion module together with the rule pairing the two streams cannot isolate the contribution of that rule.

\subsection{Cross-Modal Attention and Fusion}

Fusion methods of this family share the premise that learned interaction between the streams improves prediction. Cross-modal attention, in which the tokens of one modality attend to those of the other, was introduced for unaligned sequences by the Multimodal Transformer~\cite{tsai2019mult} and is used across multimodal sequence modeling and emotion recognition~\cite{praveen2022jca,kawamura2025crossmodal,deeb2025crossattention}. The exchange can be restricted to a few shared bottleneck tokens~\cite{nagrani2021mbt}, or carried by cross-modal adapters inside frozen vision transformers~\cite{lin2023lavish}. Other methods regulate what enters the fusion, through audio-guided visual gates~\cite{zhang2026sadgr} or a decomposition into shared and modality-specific components~\cite{qian2026decalign,cho2025modality}. These mechanisms differ in where the streams interact, and each keeps a learned selection of the information that enters fusion. The presence of learned cross-modal weights does not establish that predictions depend on the particular correspondences those weights encode.

\subsection{Optimal Transport in Attention and Alignment}

Transport-based attention and related assignment methods treat routing as an allocation problem governed by normalization or capacity constraints. Sinkhorn scaling solves entropic transport under prescribed marginals~\cite{cuturi2013sinkhorn,peyre2019computational}, and unbalanced formulations relax those marginals through divergence penalties~\cite{chizat2018scaling,sejourne2023uot}. Inside attention itself, Sparse Sinkhorn Attention learns a sorting of key blocks and Sinkformers replace the softmax with Sinkhorn normalization~\cite{tay2020sparsesinkhorn,sander2022sinkformers}. Token-to-expert routing makes the same choice of totals: BASE layers give each token one expert under equal token counts per expert~\cite{lewis2021base}, whereas expert-choice routing fixes the tokens per expert and leaves the experts per token free~\cite{zhou2022expertchoice}. These mechanisms differ in which allocation totals are prescribed, relaxed, or left free. For the unregularized problem, the independent coupling is optimal exactly when the cost is separable~\cite{peyre2026otml}.

Transport-based multimodal alignment assumes that useful cross-modal correspondence can be represented by a transport plan. MM-Align~\cite{han2022mmalign}, AlignMamba~\cite{li2025alignmamba}, Att-Sinkhorn~\cite{ma2023attsinkhorn} and LAVCap~\cite{rho2025lavcap} align modality representations this way. The same mechanism can carry linguistic cues: GRACE aligns emotion-aware text with video segments through entropic transport~\cite{liu2026grace}, and transport plans fuse text with several visual views in multimodal named entity recognition~\cite{wu2026dgot}. The transport formulation specifies how to construct correspondences, and does not establish their contribution to prediction.

\subsection{Token Mixing and Cross-Modal Interactions}

Input-independent token mixers can preserve much of the performance of attention-based architectures. MetaFormer~\cite{yu2022metaformer} replaces attention with average pooling over neighboring tokens and remains competitive, while FNet~\cite{leethorp2022fnet} replaces self-attention with a Fourier transform at a moderate loss of accuracy. At the level of multimodal predictions, Hessel and Lee~\cite{hessel2020emap} project a trained model's predictions onto the closest function that is additive in the modalities, removing cross-modal interactions, and find that the projection often performs as well as the model. Liang et al.~\cite{liang2023pid} instead decompose the information that two modalities carry about a label into redundant, unique, and synergistic parts.

Our study examines the routing inside the fusion module rather than the predictions of the model as a whole. One shared module supplies every plan variant at a matched parameter count, so a comparison between variants does not change the module and the routing rule together. Cross-modal attention is one of those variants, the row softmax of the descriptor similarities (\Cref{sec:rules}), rather than a competing architecture.
\section{Our Approach}
\label{sec:method}

Five comparisons separate the contribution of the transport plan from the contribution of the module around it. Two train variants differing only in the rule that forms the plan: SCR against the retrained backbone, and SCR against an individually trained, parameter-matched pooled variant. Three act on the computed plan of a trained model without retraining, replacing it with uniform weights, permuting its rows, or replacing it with the product of its own marginals. SCR supplies a shared implementation for these comparisons; the design does not depend on SCR outperforming pooling.

\subsection{Individually Trained Comparisons}

The baseline of each training comparison is the backbone without the module, retrained under the same conditions as the corresponding module variants. Comparing SCR with that baseline changes capacity and fusion architecture together. The pooled variant retains the module architecture, projections, gates, and parameter count, and uses the product of the frame and slot measures as the plan. SCR and the pooled variant are trained individually, with the respective routing rule present during both training and inference. Their difference therefore evaluates the learned configurations under those rules, rather than the effect of replacing a plan after training.

\subsection{Interventions Without Retraining}

Uniform replacement substitutes $\mathbf{a}\mathbf{b}^{\top}$ for the computed plan of a trained SCR model during evaluation, without retraining. The frame and slot measures are defined in~\eqref{eq:measures}, so uniformity concerns the slots admitted by the padding mask. The replacement plan enters the existing routing updates, and the projections and gates are not retrained. The third comparison applies this replacement in every block carrying the module; confining the replacement to earlier blocks or to the final block tests whether a change depends on where the plan is replaced (\Cref{tab:inference-plan}).

Row permutation reassigns the rows of the computed plan to frame indices, also without retraining. For each clip and each selected block, a sampled permutation $\sigma$ produces
\begin{equation*}
\pi^{\mathrm{perm}}_{jk}=\pi_{\sigma(j),k}.
\end{equation*}
The operation acts on the plan after computation, leaving the frame descriptors and audio-slot descriptors in their original order. The fourth comparison applies row permutation in every block carrying the module. This operation differs from permuting audio inputs or encoded audio slots, which would change the inputs to plan computation. The original-plan evaluations agree between the intervention runs, allowing the permutation-minus-uniform contrast to be paired on the same trained models and folds.

By construction, row permutation preserves the column masses exactly and preserves the multiset of row masses exactly. Each frame receives a permuted row mass rather than necessarily retaining the original mass, so agreement with the uniform frame target remains limited by the solver residual. The reported bound $\RowResidMax{}$ concerns the largest aggregated block-mean relative row residual of the diagnosed development models, rather than a guarantee for every clip (\Cref{sec:residuals}). A non-identity permutation can retain some frame assignments, and similar plan rows make reassignment a weak perturbation. The intervention therefore disrupts the original assignment without guaranteeing removal of every useful correspondence.

Actual-marginal product replacement removes the frame-specific component instead of reassigning it. For each clip and each selected block, the computed plan supplies its own row masses $m_j$, column masses $c_k$ and total mass $S=\sum_{j,k}\pi_{jk}$, and the replacement is
\begin{equation*}
\pi^{\mathrm{prod}}_{jk}=\frac{m_j c_k}{S}.
\end{equation*}
Every row of $\boldsymbol{\pi}^{\mathrm{prod}}$ normalizes to the same mixture $\mathbf{c}/S$, so no frame keeps a mixture of its own and similar rows give no partial protection. The construction preserves each measured row mass and each measured column mass rather than substituting the target marginals, which separates this intervention from uniform replacement: the frame mass factors $\rho_j$ and the slot mass factors $\sigma_k$ of~\eqref{eq:a2v} and~\eqref{eq:v2a} are unchanged, and the identity $\omega_j=m_j$ of~\eqref{eq:v2a} leaves the vector returned to the audio stream unchanged at fixed incoming descriptors. Changes in the audio-to-video direction in earlier blocks still alter the descriptors that later blocks receive, so a complete intervened pass does not leave the audio stream unchanged. The shared mixture $\mathbf{c}/S$ follows the slot usage of the computed plan and need not equal $\mathbf{b}$, so frame independence here is not uniformity. Fewer changed properties of the plan do not imply a smaller change in UAR.

\subsection{Shared Routing Module}

\begin{figure}[t]
\centering
\includegraphics{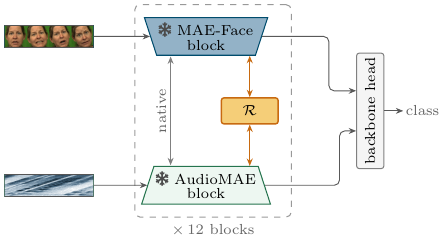}
\caption{Placement of SCR ($\mathcal{R}$). Snowflakes mark the frozen encoders.}
\label{fig:overview}
\end{figure}

SCR is attached to each transformer block of a published frozen-encoder audio-visual backbone (\Cref{fig:overview}). A transport plan between the frames and the audio time slots routes audio to each frame, and a video vector returns to the audio stream. Each transformer block holds separate module parameters. Training updates the added modules and the classification head, while the encoder weights remain frozen apart from MMA-DFER's position embeddings. The backbone dimensions and the solver are specified in \Cref{sec:module,sec:residuals}.

\textbf{Native cross-modal exchange.} Both backbones already exchange information in both directions inside every transformer block, and SCR supplements that exchange rather than replacing it. MMA-DFER projects both streams into a shared latent space and adds the mean of the audio latents to every frame, so the audio summary is shared across frames; the mean of the video latents returns to every audio token. ABVnet adds each adapted audio token to the image token at the same index in every frame, and returns the adapted image tokens averaged across frames at each index. Matching token indices pairs positions within the two sequences and carries no correspondence between frames and audio time. Both mechanisms stay in place in the baseline and in every module comparison, and the implementation is given in \Cref{sec:hosts}.

\textbf{Descriptors and cost.} The descriptor $\mathbf{v}_j$ of frame $j$ is its class-token latent. The descriptor $\mathbf{a}_k$ of slot $k$ averages the patch latents of its time column over frequency; each slot covers 0.16~s. After a modality-specific LayerNorm and $L^2$ normalization, written $\hat{\mathbf{v}}_j$ and $\hat{\mathbf{a}}_k$, the cost is the cosine distance
\begin{equation}
C_{jk} = 1 - \langle \hat{\mathbf{v}}_j, \hat{\mathbf{a}}_k \rangle \in [0,2].
\label{eq:cost}
\end{equation}

The $L^2$ normalization removes the magnitude differences between the LayerNorm outputs, so the cost depends on the directions of the two descriptors.

The padding mask excludes slots carrying only padding from the slot measure, unless no slot carries audio. A clip supplies $T$ face frames and a fixed audio window divided into $M$ time slots. Both backbones take their audio from a fixed window at the start of the clip, whereas the frames are sampled over the whole clip. Audio shorter than that window is zero-padded. A uniform measure over all slots would make every variant treat the padded slots as audio. Writing $\mathcal{K}$ for the slots that carry audio, the measures are
\begin{equation}
a_j=\frac{1}{T},\qquad
b_k=
\begin{cases}
1/|\mathcal{K}| & \text{if } k\in\mathcal{K},\\
0 & \text{otherwise}.
\end{cases}
\label{eq:measures}
\end{equation}
If $\mathcal{K}=\emptyset$, the slot measure is uniform over all $M$ slots and the time prior is not applied. Every slot sum and ratio below is evaluated on the support $b_k>0$. The mask changes which slots count as audio, but cannot restore audio beyond the fixed window.

\textbf{Audio-to-video update.} Normalizing each row of the plan gives an audio mixture for the corresponding frame. Normalizing discards how much mass the frame actually received, which the frame mass factor $\rho_j=m_j/a_j=Tm_j$ of SCR restores, with row mass $m_j=\sum_k\pi_{jk}$. The update is
\begin{equation}
\begin{gathered}
\mathbf{r}_j = \rho_j \sum_{k} \frac{\pi_{jk}}{m_j}\,\mathbf{a}_k, \\
\mathbf{V}_j \leftarrow \mathbf{V}_j + \mathbf{1}\big(\tanh(\gamma)\,\mathbf{W}\mathbf{r}_j\big)^{\top},
\end{gathered}
\label{eq:a2v}
\end{equation}
where $\mathbf{W}$ is a linear layer and $\gamma$ a scalar gate. The same projected mixture is added to every token of frame $j$. This gated projection is the routing path the module adds, and the plan decides which audio travels along it. The gate starts at zero, so the module leaves the backbone unchanged at initialization, and $\rho_j$ approaches one as the frame marginal converges to $\mathbf{a}$.

\textbf{Video-to-audio update.} The audio stream receives one vector per clip through a linear layer $\mathbf{W}'$ and a scalar gate $\gamma'$, also starting at zero. Each slot mixes the frame descriptors through its column of the plan, normalized by the column mass $c_k=\sum_j\pi_{jk}$. SCR then scales that mixture by the slot mass factor $\sigma_k=c_k/b_k$, which restores the mass the slot gave relative to $b_k$. The projected slot vectors are then averaged with the weights $\mathbf{b}$, and by linearity the projection can follow that average,
\begin{equation}
\begin{gathered}
\mathbf{s} = \sum_{j} \omega_j\,\mathbf{v}_j, \qquad
\omega_j = \sum_{k} b_k\,\sigma_k\,\frac{\pi_{jk}}{c_k}, \\
\mathbf{A} \leftarrow \mathbf{A} + \mathbf{1}\big(\tanh(\gamma')\,\mathbf{W}'\mathbf{s}\big)^{\top},
\end{gathered}
\label{eq:v2a}
\end{equation}
where $\omega_j$ is the weight of frame $j$ in the returned vector. Under SCR, substituting $\sigma_k=c_k/b_k$ into~\eqref{eq:v2a} gives $\omega_j=\sum_k \pi_{jk}=m_j$ exactly on the slots with $b_k>0$. For fixed frame descriptors the returned vector therefore depends on the plan only through its row marginal: two plans with the same row masses give the same $\mathbf{s}$. Under exact balanced constraints $\mathbf{s}=\sum_j a_j\mathbf{v}_j$ for any cost, with no separability assumption, while at the finite penalty and ten iterations used here the row masses can still depend on the cost. In this direction the plan only reweights the frames that form one vector, which is then added to every audio token, so the frame-specific slot mixtures of \Cref{sec:mechanism} concern the audio-to-video direction.

\subsection{Plan Variants}
\label{sec:rules}

The variants retain the shared module and parameter count while changing the plan rule. SCR and the unbalanced variant use the same entropic unbalanced optimal transport (UOT) formulation~\cite{cuturi2013sinkhorn,chizat2018scaling,peyre2019computational}, given in~\eqref{eq:uot}: SCR targets the balanced problem through a large finite marginal penalty, whereas UOT relaxes the marginals more strongly. A finite iteration budget approximates the penalized optimum rather than enforcing the marginals exactly.

That problem penalizes the departure of the plan from the product of the two measures, and the departure of its marginals from $\mathbf{a}$ and $\mathbf{b}$:
\begin{align}
\min_{\boldsymbol{\pi}\ge 0}\ & \langle \mathbf{C},\boldsymbol{\pi}\rangle
+ \varepsilon\,\mathrm{KL}(\boldsymbol{\pi}\,\|\,\mathbf{a}\mathbf{b}^{\top}) \nonumber\\
& + \tau\,\mathrm{KL}(\boldsymbol{\pi}\mathbf{1}\,\|\,\mathbf{a})
+ \tau\,\mathrm{KL}(\boldsymbol{\pi}^{\top}\mathbf{1}\,\|\,\mathbf{b}).
\label{eq:uot}
\end{align}
Here $\mathrm{KL}$ is the generalized Kullback--Leibler (KL) divergence
$\mathrm{KL}(\mathbf{p}\,\|\,\mathbf{q})=\sum_i[p_i\log(p_i/q_i)-p_i+q_i]$, summed over the entries of a matrix
argument, with $0\log 0=0$.

The variants differ in which marginals they constrain. SCR approaches both target marginals, UOT penalizes departures from both through KL terms, the row softmax fixes the frame marginal and leaves the slot marginal free, the column softmax does the reverse, and the pooled variant fixes both. The row softmax is cross-modal attention from frames to audio slots~\cite{tsai2019mult}. The pooled variant sets $\boldsymbol{\pi}=\mathbf{a}\mathbf{b}^{\top}$ independently of the cost. SCR and UOT apply the mass factors of the shared updates, while the softmax and pooled variants use unit mass factors. Training settings are given in \Cref{sec:training}.
\section{Comparisons and Results}
\label{sec:expr}

The comparisons in ABVnet with DFEW test whether the module-level gain can be attributed to the transport plan (\Cref{tab:comparisons}). The paired contrast between row permutation and uniform replacement carries the argument, since a resolved difference between them narrows the interpretation of uniform-replacement damage without treating an unresolved difference as zero. The training comparisons and the interventions are not successive components to subtract from the baseline gain.

\begin{table}[t]
\centering
\caption{ABVnet on DFEW: paired mean UAR differences over \SwapFoldCount{} folds, with
\SwapCILevel{} paired-fold $t$ confidence intervals (CI). Each replacement affects every block
and is compared with the original plan; the final row contrasts row permutation with uniform
replacement on the same models and folds. Under the \FoldFactor{} widening, only the gain over the baseline remains resolved (\Cref{sec:foldsens}).}
\label{tab:comparisons}
\setlength{\tabcolsep}{6pt}
\fitwidth{%
\begin{tabular}{@{}p{0.48\columnwidth}rc@{}}
\toprule
Comparison
& \begin{tabular}[b]{@{}c@{}}$\Delta$UAR\\(points)\end{tabular}
& CI \\
\midrule
\multicolumn{3}{@{}l}{\textit{Individually trained}} \\
\addlinespace
\hspace{1em}SCR minus retrained baseline
& $+\gScrDfew{}$
& $\gScrDfewCI{}$ \\
\hspace{1em}SCR minus pooled variant
& $+\mScrDfew{}$
& $\mScrDfewCI{}$ \\
\midrule
\multicolumn{3}{@{}l}{\textit{Without retraining}} \\
\addlinespace
\hspace{1em}Uniform replacement
& $-\SwapAbvDfewAllMag{}$
& $\SwapAbvDfewAllCI{}$ \\
\hspace{1em}Row permutation
& $\PermAbvDfew{}$
& $\PermAbvDfewCI{}$ \\
\hspace{1em}Product replacement
& $-\ProdAbvDfewAllMag{}$
& $\ProdAbvDfewAllCI{}$ \\
\addlinespace
\hspace{1em}Permutation minus uniform
& $\PermMinusSwap{}$
& $\PermMinusSwapCI{}$ \\
\bottomrule
\end{tabular}}
\end{table}

\subsection{Evaluation Protocol}
\label{sec:setup}

MMA-DFER and ABVnet follow the official \SwapFoldCount{}-fold cross-validation splits for the in-the-wild datasets DFEW~\cite{jiang2020dfew} and MAFW~\cite{liu2022mafw}. The additional MMA-DFER evaluation on studio-recorded CREMA-D~\cite{cao2014crema} uses \SwapFoldCount{} actor-disjoint folds constructed for this study, to examine routing in a controlled setting with frontal faces and continuous speech; the fold construction is given in \Cref{sec:app-repro}.

Both backbones follow the training recipes of~\cite{chumachenko2024mmadfer,yang2026abvnet} and are evaluated at the last epoch, with the training settings given in \Cref{sec:training}. By default, the module is inserted in every transformer block at temperature $\varepsilon=\epsdef$. Following prior work on these datasets, we report unweighted average recall (UAR), the mean recall across classes, and weighted average recall (WAR), the overall accuracy, as means over the \SwapFoldCount{} folds.

The development protocol holds out a tenth of each training split.
Decisions about the module, its variants and their hyperparameters were made under the
development protocol; the test protocol trains on the whole split and evaluates once, and the
interventions of \Cref{sec:inference-plan} re-evaluate those same trained weights without
further training. The focal
setting itself was chosen after the broader results were available (\Cref{sec:limitations}), so
the development protocol does not cover that choice. \Cref{tab:dev} gives the development means.

The estimand is the mean difference in UAR over the \SwapFoldCount{} folds of one dataset, for one training run of each configuration under one fixed recipe, with the folds as the unit of analysis. Each difference is taken within a fold, from the unrounded values rather than the rounded means shown in the tables. A configuration's standard deviation over folds, the quantity most published work reports, describes the dispersion of its scores; an interval on a paired difference instead addresses uncertainty in that difference, and any additive fold effect shared by the two configurations cancels within a fold. The mean of the per-fold differences therefore carries a nominal \SwapCILevel{} confidence interval from the paired-fold $t$ distribution, and a difference whose interval excludes zero is called resolved. The intervals describe variation across folds only, and pairing leaves the dependence between fold differences that overlapping training splits create (\Cref{sec:foldsens,sec:limitations}).

\subsection{Module Gain and the Pooled Variant}
\label{sec:confirm}

\begin{table*}[!t]
\centering
\caption{Test results, \SwapFoldCount{}-fold means. Baseline: the retrained backbone without
the module; the individually trained variants share architecture and parameter count.
$\Delta$UAR: the mean of the unrounded paired fold differences, with \SwapCILevel{} interval
half-widths \ConfirmCIHalfLo{} to \ConfirmCIHalfHi{} UAR. ABVnet SCR paired intervals on DFEW
and MAFW: to the baseline, \gScrDfewCI{} and \gScrMafwCI{}; to the pooled variant,
\mScrDfewCI{} and \mScrMafwCI{}. Fold standard deviations are given in \Cref{tab:foldsd}.}
\label{tab:confirm}
\setlength{\tabcolsep}{4pt}
\fitwidth{%
\begin{tabular}{@{}llccccccc@{}}
\toprule
 & & \multicolumn{4}{c}{UAR / WAR} & \multicolumn{3}{c}{$\Delta$UAR to baseline} \\
\cmidrule(lr){3-6}\cmidrule(l){7-9}
Backbone & Dataset & Baseline & SCR & Pooled & Column softmax & SCR & Pooled & Column softmax \\
\midrule
\multirow{3}{*}{MMA-DFER} & DFEW & \MmaDfewBase & \MmaDfewScr & \MmaDfewPool & 65.62 / 76.38 & $-0.08$ & $+0.06$ & $-0.27$ \\
 & MAFW & \MmaMafwBase & \MmaMafwScr & \MmaMafwPool & 43.59 / 56.71 & $+0.11$ & $-0.20$ & $+0.11$ \\
 & CREMA-D & 80.35 / 80.28 & 80.44 / 80.39 & 80.47 / 80.41 & 80.43 / 80.42 & $+0.09$ & $+0.12$ & $+0.08$ \\
\midrule
\multirow{2}{*}{ABVnet} & DFEW & \AbvDfewBase & \AbvDfewScr & \AbvDfewPool & 66.42 / 77.19 & $+\gScrDfew$ & $+\gPoolDfew$ & $-0.26$ \\
 & MAFW & \AbvMafwBase & \AbvMafwScr & \AbvMafwPool & 43.80 / 57.66 & $+\gScrMafw$ & $+\gPoolMafw$ & $+0.00$ \\
\bottomrule
\end{tabular}}
\end{table*}

\begin{table*}[!t]
\centering
\caption{Development results: the ABVnet baseline on DFEW exceeds both SCR and the pooled
variant, reversing the SCR-against-baseline comparison of \Cref{tab:confirm}. No plan variant
shown here has a resolved advantage over the pooled variant. Values are mean UAR over
\SwapFoldCount{} folds and paired $\Delta$UAR against the pooled variant, from unrounded fold
values.}
\label{tab:dev}
\fitwidth{%
\begin{tabular}{@{}llccccccc@{}}
\toprule
 & & \multicolumn{4}{c}{UAR} & \multicolumn{3}{c}{$\Delta$UAR to pooled} \\
\cmidrule(lr){3-6}\cmidrule(l){7-9}
Backbone & Dataset & Baseline & SCR & Pooled & Column softmax & Baseline & SCR & Column softmax \\
\midrule
MMA-DFER & DFEW & 66.45 & 66.74 & 67.27 & 67.45 & $-0.82^{\dagger}$ (2/5) & $-0.54$ (2/5) & $+0.17$ (2/5) \\
MMA-DFER & MAFW & 49.05 & 49.08 & 48.89 & 49.08 & $+0.17$ (4/5) & $+0.19$ (3/5) & $+0.19$ (3/5) \\
MMA-DFER & CREMA-D & 75.39 & 75.35 & 75.54 & 74.73 & $-0.15$ (2/5) & $-0.19$ (1/5) & $-0.81$ (1/5) \\
\midrule
ABVnet & DFEW & 67.52 & 66.17 & 66.27 & 66.90 & $+1.25^{\dagger}$ (5/5) & $-0.10$ (2/5) & $+0.63$ (2/5) \\
\bottomrule
\end{tabular}}
\par\smallskip
{\footnotesize\raggedright
Parentheses: folds with higher UAR than the pooled variant.
$\dagger$: the paired 90\% interval excludes zero.\par}
\end{table*}

\begin{figure*}[!t]
\centering
\includegraphics[width=\textwidth]{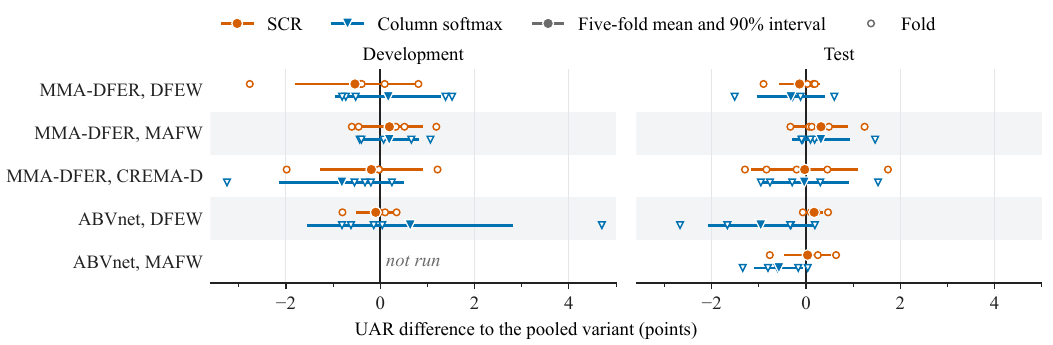}
\caption{No plan variant shows a resolved advantage over the pooled variant. Paired
UAR difference to the pooled variant for SCR and the column softmax, over the
\SwapFoldCount{} folds of each setting under both protocols; filled markers are the
fold means, bars the paired \SwapCILevel{} intervals, hollow markers the individual
folds. The one interval excluding zero is the ABVnet MAFW column softmax, a
disadvantage.}
\label{fig:variants}
\end{figure*}

SCR exceeds the retrained ABVnet baseline by \gScrDfew{} UAR points on DFEW under the test protocol, a resolved difference (\Cref{tab:comparisons}). Reproduction gaps restrict this gain to the internal comparison, and the same SCR folds give an unresolved difference of \ScrFpDelta{} UAR points, \ScrFpCI{}, against the single-precision baseline (\Cref{sec:app-repro}). Under the development protocol, the ABVnet comparison with the baseline reverses on DFEW: the baseline exceeds SCR by \DevAbvBaseScr{} UAR points and also exceeds the pooled variant, with both differences positive in every fold (\Cref{tab:dev}). The test and development estimates concern different training protocols. The gain is spread unevenly over the classes, with \PerClassResolved{} resolved per-class difference among \PerClassCount{} (\Cref{tab:perclass}). The test result supports a benefit from the complete module over this retrained baseline; attribution to the transport plan and a controlled improvement over the published model remain unestablished.

For ABVnet on DFEW, SCR minus the individually trained, parameter-matched pooled variant is \mScrDfew{} UAR points, an unresolved difference (\Cref{tab:comparisons}). An unresolved difference does not establish equivalence~\cite{schuirmann1987tost,lakens2017equivalence}. Across the default-placement SCR-against-pooled comparison family, the largest upper interval endpoint is bounded above by \ScrPoolUpperMax{} UAR points. Under the paired-fold model, the minimum detectable effect for a two-sided test at the \SwapCILevel{} confidence level is bounded above by \MDEScrPool{} UAR points at power \ScrPoolPower{}, using the family's median standard deviation of paired fold differences.

Under the test protocol, the pooled variant has higher mean UAR than the retrained ABVnet baseline on both DFEW and MAFW, although only the DFEW difference is resolved (\Cref{tab:confirm}). On MAFW, SCR also has higher mean UAR than the pooled variant, but the difference is unresolved. On MMA-DFER, the absolute mean difference between SCR and the retrained baseline is at most \MmaBound{} UAR points across DFEW, MAFW and CREMA-D. Under the development protocol, the pooled variant gains \DevMmaPoolBase{} UAR points over the MMA-DFER baseline on DFEW, the only resolved difference among the MMA-DFER development comparisons with the baseline. That gain is consistent with a benefit from adding the pooled module beside the native exchange (\Cref{sec:hosts}), although the corresponding test difference is unresolved.

Neither SCR nor the column softmax has a resolved UAR advantage over the pooled variant in the evaluated default-placement settings (\Cref{fig:variants,tab:dev,tab:confirm}). Under the test protocol, the column softmax has lower mean UAR than the pooled variant on ABVnet, by \PlanColPoolDfew{} UAR points on DFEW and \PlanColPoolMafw{} on MAFW; only the MAFW difference is resolved.

\subsection{Paired Plan Interventions}
\label{sec:inference-plan}

\begin{table*}[t]
\centering
\caption{Two plan replacements in trained SCR models on the test folds, holding the weights
and the other blocks fixed; both rules are defined in \Cref{sec:method}. Blocks: the intervened
blocks. $\Delta$UAR: the replacement minus the same model's original-plan UAR in the same pass.
Chg.: the share of test clips whose predicted class differs. The last pair is the paired
contrast between the two replacements, formed fold by fold against the same unmodified model.
Differences average \SwapFoldCount{} folds; every interval is a \SwapCILevel{} paired
confidence interval.}
\label{tab:inference-plan}
\footnotesize
\setlength{\tabcolsep}{2pt}
\fitwidth{%
\begin{tabular}{@{}lll cc cc cc@{}}
\toprule
 & & & \multicolumn{2}{c}{Uniform plan} & \multicolumn{2}{c}{Row permutation}
 & \multicolumn{2}{c}{Permutation $-$ uniform} \\
\cmidrule(lr){4-5} \cmidrule(lr){6-7} \cmidrule(l){8-9}
Backbone & Dataset & Blocks & $\Delta$UAR & \begin{tabular}[b]{@{}c@{}}Chg.\\(\%)\end{tabular}
 & $\Delta$UAR & \begin{tabular}[b]{@{}c@{}}Chg.\\(\%)\end{tabular}
 & $\Delta$UAR & CI \\
\midrule
MMA-DFER  & DFEW     & 1--12  & $-0.01$ & 0.30 & $+0.04$ & 0.32 & $+0.04$ & $[-0.01, +0.10]$ \\
          &          & 1--11  & $0.00$ & 0.26 & $+0.03$ & 0.27 & $+0.02$ & $[-0.04, +0.08]$ \\
          &          & 12     & $-0.02$ & 0.12 & $-0.02$ & 0.14 & $0.00$ & $[-0.06, +0.07]$ \\
\midrule
MMA-DFER  & MAFW     & 1--12  & $0.00$ & 0.84 & $+0.01$ & 0.97 & $+0.01$ & $[-0.13, +0.14]$ \\
          &          & 1--11  & $+0.12$ & 0.69 & $+0.04$ & 0.64 & $-0.07$ & $[-0.15, +0.00]$ \\
          &          & 12     & $0.00$ & 0.39 & $0.00$ & 0.39 & $0.00$ & $[-0.10, +0.10]$ \\
\midrule
MMA-DFER  & CREMA-D  & 1--12  & $-0.06$ & 0.51 & $-0.09$ & 0.57 & $-0.02$ & $[-0.12, +0.07]$ \\
          &          & 1--11  & $0.00$ & 0.27 & $-0.05$ & 0.28 & $-0.05$ & $[-0.10, +0.00]$ \\
          &          & 12     & $-0.06$ & 0.38 & $-0.02$ & 0.41 & $+0.04$ & $[+0.01, +0.08]$ \\
\midrule
ABVnet    & DFEW     & 1--12  & $-0.72^{\dagger}$ & 2.92 & $-0.20$ & 2.58 & $+0.51$ & $[+0.15, +0.88]$ \\
          &          & 1--11  & $-0.54$ & 1.80 & $-0.18$ & 1.57 & $+0.35$ & $[+0.10, +0.61]$ \\
          &          & 12     & $-0.40$ & 1.58 & $-0.29$ & 1.68 & $+0.11$ & $[-0.14, +0.37]$ \\
\midrule
ABVnet    & MAFW     & 1--12  & $-0.14$ & 3.05 & $+0.03$ & 2.89 & $+0.17$ & $[-0.20, +0.53]$ \\
          &          & 1--11  & $-0.12$ & 2.04 & $+0.01$ & 2.03 & $+0.13$ & $[-0.06, +0.32]$ \\
          &          & 12     & $+0.04$ & 1.84 & $+0.09$ & 1.81 & $+0.06$ & $[-0.13, +0.24]$ \\
\bottomrule
\end{tabular}}
\par\smallskip
{\footnotesize\raggedright
$\dagger$: the paired \SwapCILevel{} interval of that difference excludes zero; no interval is
printed for the two replacements.\par}
\end{table*}

For ABVnet on DFEW, replacing the plan in every block with the uniform plan without retraining reduces UAR by \SwapAbvDfewAllMag{} points, a resolved difference (\Cref{tab:inference-plan}). Replacement confined to the final block gives an unresolved difference, leaving localization of the all-block reduction unestablished. Uniform replacement changes the mass distribution of the plan as well as the frame--slot correspondence, so the reduction does not establish a benefit from the original correspondence.

Row permutation gives an unresolved difference from the original plan, leaving the benefit of the original frame--slot correspondence unestablished. Paired on the same trained models and folds, permutation minus uniform replacement is \PermMinusSwap{} UAR points, a resolved difference indicating less damage from row permutation (\Cref{tab:inference-plan}). Neither the uniform-replacement result nor the paired contrast retains resolution under the \FoldFactor{} fold-dependence widening (\Cref{sec:foldsens}). Row permutation preserves column masses and the multiset of row masses, but attribution of the smaller loss to a particular preserved property remains unestablished.

Actual-marginal product replacement in every block reduces UAR by \ProdAbvDfewAllMag{} points, \ProdAbvDfewAllCI{}, a resolved difference under the reported convention. The same interval widens to \ProdAbvDfewAllWideCI{} and includes zero under the \FoldFactor{} fold-dependence widening (\Cref{sec:foldsens}). Both measured marginals and both mass factors survive this replacement, so the reduction concerns the frame-specific slot mixtures alone, which the uniform-replacement reduction does not. Confining the replacement to blocks 1--11 or to the final block leaves localization unestablished: the two restricted reductions exceed the all-block reduction when added, and the final-block difference changes sign across folds. Attribution of the module gain over the retrained baseline to the transport plan remains unestablished, since replacing a plan after training does not decompose that gain.

\subsection{Evidence Beyond the Focal Setting}

The remaining settings are ABVnet with MAFW and MMA-DFER with DFEW, MAFW, and CREMA-D. For interventions applied in every block, the uniform-replacement differences, the row-permutation differences, the product-replacement differences and the paired permutation-minus-uniform contrasts are unresolved; \Cref{tab:inference-plan} gives all but the product replacement. Product replacement changes UAR by at most \ProdOtherBound{} points in absolute value across these four settings, so the resolved reduction of \Cref{sec:inference-plan} holds in \ProdResolvedCount{} of the \ProdSettingCount{} settings measured. Uniform replacement changes UAR by at most \SwapMmaAllBound{} points on the three MMA-DFER datasets, and the absolute row-permutation differences are bounded by \PermOtherMax{} UAR in every setting other than ABVnet with DFEW. Over the narrower block groups, the paired contrast excludes zero for ABVnet with DFEW over blocks 1--11 and for MMA-DFER with CREMA-D over block 12, whereas no replacement outside ABVnet with DFEW excludes zero.

Across the default-placement family of individually trained comparisons, the SCR advantage over pooling remains unresolved, and the absolute mean differences are bounded by \PlanScrPoolBound{} UAR (\Cref{tab:confirm,tab:dev,fig:variants}). This bound does not cover the restricted-placement experiment: with the module restricted to the last three blocks at $\varepsilon=0.02$, SCR sits below the pooled variant by \AxesScrLate{} UAR, \AxesScrLateCI{}, a resolved disadvantage, and is higher in none of the five folds (\Cref{tab:axes}).

\subsection{Fold Sensitivity}
\label{sec:foldsens}

The count below covers the individually trained comparisons in
\Cref{tab:confirm,tab:dev} and, in \Cref{tab:inference-plan}, the uniform-replacement
differences, the row-permutation differences, and the paired contrasts between them, together
with the product-replacement differences for replacement in every block
(\Cref{sec:inference-plan}). The
design-axis comparisons in \Cref{tab:axes} and the per-class differences in
\Cref{tab:perclass} are counted separately.

\textbf{Multiplicity.} Of \FoldTotal{} differences carrying an interval at the
\SwapCILevel{} level, \FoldResolved{} have intervals excluding zero. Under a global null and
nominal coverage, about \FoldExpected{} intervals would exclude zero by chance, so the resolved
differences are not individually protected against selection. No multiplicity adjustment is applied, and the
two claims of \Cref{sec:intro} were fixed before this count rather than chosen from it.

\textbf{Leave-one-fold-out.} Recomputing each difference on the five four-fold subsets
separates the differences that rest on the whole fold set from those resting on particular
folds. The increase of SCR over the retrained ABVnet baseline on DFEW stays resolved in all
\LooGainKept{} subsets. The paired permutation-minus-uniform contrast loses resolution in
\LooContrastLost{} of the five, and uniform replacement in \LooSwapLost{}. Omitting a fold changes both the sample and the
interval calculation, so a loss of resolution cannot be attributed to the omitted fold alone.

\textbf{Fold dependence.} Widening every reported half-width by the
corrected-resampling factor \FoldFactor{} for five folds~\cite{nadeau2003inference}, as
\Cref{sec:limitations} describes,
leaves \FoldKept{} differences excluding zero: the ABVnet test increase over the retrained baseline on
DFEW, and the two ABVnet development comparisons on DFEW in which the baseline exceeds SCR and
the pooled variant. The widened intervals include zero for uniform replacement, for product replacement, for the column
softmax on ABVnet with MAFW, and for the pooled variant against the baseline both on MMA-DFER
development data and on ABVnet with DFEW test data. The paired permutation-minus-uniform
contrast widens to \PermMinusSwapWideCI{} and also includes zero, as do the two remaining
resolved contrasts in \Cref{tab:inference-plan}.

\subsection{Cost Interaction and Routing}
\label{sec:mechanism}

For the routing rules in \Cref{sec:rules}, a normalized slot mixture can vary across frames only if the cost couples frames and slots. The decomposition against the frame and slot measures is
\begin{equation}
\begin{gathered}
C_{jk}=\mu+u_j+w_k+I_{jk},\\
\mu=\textstyle\sum_{j,k}a_jb_kC_{jk},\\
u_j=\textstyle\sum_k b_kC_{jk}-\mu,\qquad
w_k=\textstyle\sum_j a_jC_{jk}-\mu.
\end{gathered}
\label{eq:decomp}
\end{equation}
The interaction $\mathbf{I}$ has vanishing $\mathbf{b}$-weighted row means and $\mathbf{a}$-weighted column means. The frame and slot terms give the average cost of a frame and of a slot relative to $\mu$, and $\mathbf{I}$ carries the remaining dependence between the indices. A separable cost has $\mathbf{I}=\mathbf{0}$ on the supported slots.

\begin{corollary}[Pooling under a separable cost, following~\cite{sander2022sinkformers,peyre2026otml}]
\label{cor:separable}
Adding frame-only or slot-only terms to the cost leaves the converged balanced entropic plan unchanged. For a separable cost, the balanced plan is therefore $\mathbf{a}\mathbf{b}^{\top}$. The balanced plan and the column softmax have normalized row weights $\mathbf{b}$, and the row softmax has normalized row weights
\begin{equation*}
\mathbf{p}=\mathrm{softmax}(-\mathbf{w}/\varepsilon+\log\mathbf{b}),
\end{equation*}
shared by all frames. Under the update conventions of \Cref{sec:rules}, every frame receives the pooled audio $\bar{\mathbf{a}}=\sum_k b_k\mathbf{a}_k$ under balanced routing and the column softmax, or $\sum_k p_k\mathbf{a}_k$ under the row softmax. At finite marginal penalty, SCR and UOT retain a factorized plan for a separable cost after every complete scaling cycle; the shared normalized slot weights are proportional to $b_k e^{-w_k/(\tau+\varepsilon)}$, while the mass factor $\rho_j$ scales the audio vector of frame $j$. Thus shared normalized mixtures do not require identical update magnitudes at finite penalty. The derivation is given in \Cref{sec:proofs}.
\end{corollary}

For measurements on trained blocks, we summarize the interaction relative to the temperature by
\begin{equation}
\kappa_{\mathrm{int}}
=\frac{1}{\varepsilon}
\Big(\textstyle\sum_{j,k}a_jb_k I_{jk}^{2}\Big)^{1/2}.
\label{eq:kappa}
\end{equation}
This diagnostic measures the weighted root-mean-square cost interaction. The entrywise bound in \Cref{prop:near} uses the largest interaction magnitude and applies to the converged balanced plan, rather than the finite-penalty, finite-iteration SCR implementation. Measured marginal residuals are reported in \Cref{sec:residuals}. A large $\kappa_{\mathrm{int}}$ leaves room for frame-specific routing without implying a gain in UAR.

We measured the cost decomposition and resulting plans of trained SCR models block by block on development clips, averaging over clips within a fold and then over folds, without further training (\Cref{tab:routing}). On MMA-DFER, normalized routing weights stay close to uniform before the final block on every evaluated dataset, with a mean slot ratio of at least \SlotRatioMin{}. The mean frame-specific update is at most \FramePartMax{} of the receiving image-token norm in those blocks. The slot ratio measures concentration of normalized routing weights, whereas the token-relative update combines variation across frames with total update magnitude. A small token-relative update therefore does not establish near-uniform weights. In the final block, the cost interaction becomes comparable to the temperature and the plan departs furthest from uniform.

Neither diagnostic measures how far the mixture of one frame departs from a mixture shared across frames. On the test models of every setting we therefore measured
\begin{equation*}
H=\operatorname{mean}_j \tfrac{1}{2}\big\|\boldsymbol{\pi}_j/m_j-\mathbf{c}/S\big\|_1,
\end{equation*}
the mean total variation between the normalized mixture of a frame and the clip-level mixture $\mathbf{c}/S$, averaged over the frames carrying mass and evaluated on \HPairs{} clip-and-block observations. $H$ is the displacement that actual-marginal product replacement applies, so the quantity measures the frame-specific structure available to that intervention; $H$ bounds neither the logits nor UAR, and row permutation admits only $2H$ as a bound on the same displacement. Median $H$ is at most \HMedianBound{} in every setting, and at least \HShareMin{} of observations sit at or below \HLevel{}. Block means increase overall with depth and are largest in the final block in every setting, without increasing monotonically; first-block means stay at or below \HFirstBound{}. On MMA-DFER the means stay at or below \HMmaEarlyBound{} through block 11 and reach \HFinalMmaLo{} to \HFinalMmaHi{} in the final block, whereas the ABVnet final-block means are \HFinalAbvLo{} to \HFinalAbvHi{}. These observations cluster within clips and trained models, so the counts describe coverage rather than supplying folds for a recognition comparison.

\begin{table}[t]
\newcommand{\RoutingMmaDfewEarly}{1--9 & 0.00--0.04 & 0.2--2.4 & 0.999--1.000 & 0.00--0.06}
\newcommand{\RoutingMmaDfewTen}{10 & 0.09 & 6.1 & 0.994 & 0.44}
\newcommand{\RoutingMmaDfewEleven}{11 & 0.10 & 3.2 & 0.993 & 1.14}
\newcommand{\RoutingMmaDfewFinal}{12 & 0.75 & 22.7 & 0.854 & 2.99}
\newcommand{\RoutingMmaMafwEarly}{1--9 & 0.00--0.05 & 0.2--2.9 & 0.998--1.000 & 0.00--0.13}
\newcommand{\RoutingMmaMafwTen}{10 & 0.15 & 9.6 & 0.985 & 0.77}
\newcommand{\RoutingMmaMafwEleven}{11 & 0.15 & 5.8 & 0.985 & 1.35}
\newcommand{\RoutingMmaMafwFinal}{12 & 0.91 & 28.9 & 0.819 & 2.22}
\newcommand{\RoutingMmaCremaEarly}{1--9 & 0.00--0.03 & 0.1--1.8 & 0.999--1.000 & 0.00--0.05}
\newcommand{\RoutingMmaCremaTen}{10 & 0.12 & 7.4 & 0.991 & 0.91}
\newcommand{\RoutingMmaCremaEleven}{11 & 0.13 & 3.6 & 0.989 & 1.71}
\newcommand{\RoutingMmaCremaFinal}{12 & 1.13 & 29.9 & 0.762 & 5.67}
\newcommand{\RoutingAbvDfewEarly}{1--5 & 0.00--0.09 & 0.3--5.5 & 0.990--1.000 & 0.00--0.12}
\newcommand{\RoutingAbvDfewLater}{6--11 & 0.26--0.52 & 13--24 & \AbvSlotLo--\AbvSlotHi & 0.44--1.01}
\newcommand{\RoutingAbvDfewFinal}{12 & 0.71 & 30.2 & 0.871 & 2.21}
\centering
\caption{SCR diagnostics ($\varepsilon=\epsdef$) on development clips. Values average clips within folds, then the \SwapFoldCount{} folds; grouped blocks give the range of their block means.}
\label{tab:routing}
\setlength{\tabcolsep}{3pt}
\fitwidth{%
\begin{tabular}{@{}lcccc@{}}
\toprule
Blocks & $\kappa_{\mathrm{int}}$
& \begin{tabular}[b]{@{}c@{}}Interaction\\share (\%)\textsuperscript{a}\end{tabular}
& \begin{tabular}[b]{@{}c@{}}Slot\\ratio\textsuperscript{b}\end{tabular}
& \begin{tabular}[b]{@{}c@{}}Frame-specific\\update (\%)\textsuperscript{c}\end{tabular} \\
\midrule
\multicolumn{5}{@{}l}{\itshape MMA-DFER, DFEW} \\
\RoutingMmaDfewEarly \\
\RoutingMmaDfewTen \\
\RoutingMmaDfewEleven \\
\RoutingMmaDfewFinal \\
\midrule
\multicolumn{5}{@{}l}{\itshape MMA-DFER, MAFW} \\
\RoutingMmaMafwEarly \\
\RoutingMmaMafwTen \\
\RoutingMmaMafwEleven \\
\RoutingMmaMafwFinal \\
\midrule
\multicolumn{5}{@{}l}{\itshape MMA-DFER, CREMA-D} \\
\RoutingMmaCremaEarly \\
\RoutingMmaCremaTen \\
\RoutingMmaCremaEleven \\
\RoutingMmaCremaFinal \\
\midrule
\multicolumn{5}{@{}l}{\itshape ABVnet, DFEW} \\
\RoutingAbvDfewEarly \\
\RoutingAbvDfewLater \\
\RoutingAbvDfewFinal \\
\bottomrule
\end{tabular}}
\par\smallskip
{\footnotesize\raggedright
\textsuperscript{a}Frobenius norm of the frame--slot interaction divided by that of the centered cost, over the slots with audio.\par
\textsuperscript{b}Exponential of the entropy of a normalized plan row divided by the number of slots with audio; the uniform plan attains the maximum.\par
\textsuperscript{c}Deviation of a frame's gated audio-to-video update from the mean over frames, relative to the norm of the receiving image tokens; for ABVnet, measured in the module's latent space.\par}
\end{table}

ABVnet's later nonfinal blocks have less uniform routing weights (\Cref{tab:routing}). Their block-mean $\kappa_{\mathrm{int}}$ is larger than in the corresponding MMA-DFER blocks, and their mean slot ratio falls to \AbvSlotLo{} to \AbvSlotHi{}. The mean frame-specific update nevertheless remains small through those blocks. These measurements do not identify the mechanism responsible for the small updates. Weak cost interaction is consistent with near-uniform routing in MMA-DFER and the early blocks of ABVnet, but cannot fully explain ABVnet's later-block behavior or establish a benefit from the original correspondence.
\section{Scope and Discussion}
\label{sec:discussion}

\subsection{Attribution and Interpretation}

Evidence that additive approximations preserve multimodal predictions~\cite{hessel2020emap} concerns the whole model, whereas the present analysis examines a restriction inside the routing mechanism. A separable frame--slot cost makes the normalized routing weights independent of the frame (\Cref{cor:separable}); that property does not establish that the model is additive. Gains reported for complete transport-based modules~\cite{han2022mmalign,li2025alignmamba,ma2023attsinkhorn} likewise do not isolate the contribution of the plan. Parameter-matched plan variants address that attribution problem by retaining the surrounding module and changing only the rule that forms the plan, while paired interventions examine the response of trained models to changes in the plan. Neither comparison assumes separability in other models, and the difference between interventions does not provide a causal decomposition of the replacement damage. Removing the frame-specific component of the mixtures while leaving both measured marginals in place reduces UAR in the focal setting under the reported convention, which concerns the reliance of one trained configuration on structure present at inference rather than the origin of the module gain. The measured frame-specific structure is small in every setting and largest in the final block, so these interventions act on a narrow part of the plan.

\subsection{Limitations}
\label{sec:limitations}

With one training run per configuration, the intervals describe variation across folds rather
than variation from training randomness~\cite{reimers2017reporting,bouthillier2021variance}, and
the training parts of the folds overlap, undermining the independence assumption behind the
paired-fold intervals~\cite{bengio2004crossval}; the \FoldFactor{} widening is a sensitivity
adjustment rather than an exact correction for this design.
ABVnet with DFEW motivated the follow-up after the broader results were available,
so the focal analysis is retrospective. The development comparison reverses, and
reproduction gaps restrict the gains to internal comparisons (\Cref{sec:app-repro}).

The evidence is confined to frozen encoders and within-dataset evaluation, so behavior under
fine-tuning remains unestablished. The ability of the plan interventions to detect beneficial
routing also remains unestablished: no positive control with known temporal correspondence and
controlled interaction strength was evaluated, and the unresolved comparison with the pooled
variant does not establish equivalence (\Cref{sec:confirm}).
\section{Conclusion}
\label{sec:conclusion}

Comparing a complete routing module with its backbone evaluates the routing path and the
transport plan together, so an improvement there does not identify the contribution of the
plan. This study separated the two with individually trained, parameter-matched plan variants
and with paired interventions on the computed plans of trained models, across both backbones
and all three corpora. For ABVnet with DFEW, the evidence supports a module-level improvement
over the retrained test baseline, while the SCR-minus-pooled difference remains unresolved and
attribution of the module gain to the transport plan remains unestablished. Row permutation
damages UAR less than uniform replacement, and replacing the plan with the product of its own
marginals reduces UAR while leaving both measured marginals in place; neither difference
survives the \FoldFactor{} fold-dependence widening, and the benefit of the original
frame--slot correspondence remains unestablished. Frame-specific structure in the trained plans
is small in every setting and largest in the final block, which characterizes the structure the
product replacement removes.

In future work, a positive control carrying known temporal correspondence, with
interaction strength set rather than learned, would calibrate what these interventions are able
to detect. Frame descriptors that vary more within a clip, as fine-tuning the visual encoder
would produce, would give the cost the interaction that frame-specific routing requires.

\appendix
\crefalias{section}{appendix}
\section{Routing Implementation}

\subsection{Native Cross-Modal Exchange}
\label{sec:hosts}

\Cref{sec:method} describes both exchanges; their implementation follows. For
MMA-DFER~\cite{chumachenko2024mmadfer} the update adds the mean $\bar{\mathbf{A}}$ of the audio latents to every frame,
\begin{equation}
\mathbf{V}_j \leftarrow \mathbf{V}_j + \mathbf{1}\,\bar{\mathbf{A}}^{\top} ,
\end{equation}
where $\mathbf{1}$ is a vector of ones of the matching size. The updated latents return to the encoder tokens through a projection and a zero-initialized gate.

ABVnet~\cite{yang2026abvnet} trains adapters inside both frozen encoders and exchanges information through the second stage of its bimodal fusion adapter (BFA-2), which normalizes, projects and gates each audio token before the addition.

\subsection{Backbone Dimensions and Notation}
\label{sec:module}

\Cref{tab:notation} collects the symbols used throughout.
SCR operates in a 128-dimensional space in both backbones. On MMA-DFER that space is the backbone's own. On ABVnet the module takes the same tokens as BFA-2 and maps them from 768 to 128 dimensions with a linear layer and a Gaussian error linear unit, mapping its outputs back with bias-free linear layers.

\begin{table}[t]
\centering
\caption{Notation for one routing module inside one transformer block. The same symbols apply
to every block and to both backbones; dimensions are given in \Cref{sec:module}.}
\label{tab:notation}
\setlength{\tabcolsep}{4pt}
\fitwidth{%
\begin{tabular}{@{}lp{6.1cm}@{}}
\toprule
Symbol & Description \\
\midrule
\multicolumn{2}{@{}l@{}}{\itshape Inputs and indices}\\
$T$, $M$ & frames and audio slots per clip \\
$j$, $k$ & frame index and slot index \\
$\mathbf{V}_j$, $\mathbf{A}$ & latents of frame $j$; audio latents \\
\addlinespace
\multicolumn{2}{@{}l@{}}{\itshape Descriptors and cost}\\
$\mathbf{v}_j$, $\mathbf{a}_k$ & class-token latent of frame $j$; frequency-averaged latents of slot $k$ \\
$\mathbf{C}$, $C_{jk}$ & cost between frames and slots~\eqref{eq:cost} \\
\addlinespace
\multicolumn{2}{@{}l@{}}{\itshape Transport}\\
$\mathbf{a}$, $\mathbf{b}$ & frame and slot measures~\eqref{eq:measures} \\
$\mathcal{K}$ & slots that carry audio~\eqref{eq:measures} \\
$\boldsymbol{\pi}$, $\pi_{jk}$ & transport plan, slot $k$ to frame $j$ \\
$m_j$, $c_k$ & row mass $\sum_k\pi_{jk}$; column mass $\sum_j\pi_{jk}$ \\
$\varepsilon$, $\tau$, $\theta$ & temperature; penalty weight~\eqref{eq:uot}; damping $\tau/(\tau+\varepsilon)$ \\
$\rho_j$, $\sigma_k$ & mass factors; $Tm_j$ and $c_k/b_k$ for SCR and the unbalanced variant, both one otherwise \\
\addlinespace
\multicolumn{2}{@{}l@{}}{\itshape Module updates}\\
$\mathbf{r}_j$ & audio routed to frame $j$~\eqref{eq:a2v} \\
$\bar{\mathbf{a}}$ & pooled audio, $\sum_k b_k\mathbf{a}_k$ \\
$\mathbf{s}$, $\omega_j$ & returned video vector; frame weights~\eqref{eq:v2a} \\
\addlinespace
\multicolumn{2}{@{}l@{}}{\itshape Cost decomposition and diagnostic}\\
$\mu$, $u_j$, $w_k$, $I_{jk}$ & constant, frame, slot and interaction terms~\eqref{eq:decomp} \\
$\kappa_{\mathrm{int}}$ & interaction relative to the temperature~\eqref{eq:kappa} \\
\bottomrule
\end{tabular}}
\end{table}

\subsection{Solver and Forward Pass}
\label{sec:residuals}

The row softmax scores are the cosine similarities of the normalized descriptors, with no learned query and key maps.

At the default temperature and $\tau=10^{6}$ the damping $\theta=\tau/(\tau+\varepsilon)$~\cite{chizat2018scaling} is within $5\times10^{-8}$ of one, so SCR targets the balanced problem, whereas $\tau=1$ gives $\theta\approx0.95$ and leaves both marginals relaxed. Ten iterations approximate the penalized optimum rather than enforcing the marginals exactly; the measured residuals are reported below. Both take the ten iterations of \Cref{alg:scr} and apply the mass factors, whereas the two softmax variants and the pooled variant set both factors to one.

For every variant but the row softmax, whose column masses are free, the frame weights of~\eqref{eq:v2a} satisfy $\omega_j=m_j$.

The pooled variant sets $\boldsymbol{\pi}=\mathbf{a}\mathbf{b}^{\top}$ ($\varepsilon\to\infty$).

\begin{algorithm}[t]
\caption{Forward pass of one module $\mathcal{R}$ in one block, with $j=1,\dots,T$ throughout
and $k$ over the slots with $b_k>0$.}
\label{alg:scr}
\renewcommand{\algorithmicrequire}{\textbf{Input:}}
\renewcommand{\algorithmicensure}{\textbf{Output:}}
\begin{algorithmic}[1]
\REQUIRE frame latents $\mathbf{V}_1,\dots,\mathbf{V}_T$, audio latents $\mathbf{A}$, slot
         mask, temperature $\varepsilon$, penalty $\tau$
\ENSURE  updated $\mathbf{V}_1,\dots,\mathbf{V}_T$ and $\mathbf{A}$
\STATE $a_j \leftarrow 1/T$
\IF{$\mathcal{K}=\emptyset$}
  \STATE $b_k \leftarrow 1/M$ for every $k$, and the time prior is not applied
\ELSE
  \STATE $b_k \leftarrow 1/|\mathcal{K}|$ \hfill\COMMENT{\eqref{eq:measures}}
\ENDIF
\STATE $\mathbf{v}_j \leftarrow$ class-token latent of frame $j$
\STATE $\mathbf{a}_k \leftarrow$ patch latents of slot $k$ averaged over frequency
\IF{the pooled variant}
  \STATE $\boldsymbol{\pi} \leftarrow \mathbf{a}\mathbf{b}^{\top}$ \hfill\COMMENT{the cost is not formed}
\ELSE
  \STATE $\hat{\mathbf{v}}_j,\hat{\mathbf{a}}_k \leftarrow$ LayerNorm then $L^2$ normalization
         \hfill\COMMENT{the updates mix the unnormalized $\mathbf{v}_j,\mathbf{a}_k$}
  \STATE $C_{jk} \leftarrow 1-\langle\hat{\mathbf{v}}_j,\hat{\mathbf{a}}_k\rangle$
  \IF{the time prior is used}
    \STATE add the time prior \eqref{eq:timeprior} to $C_{jk}$
  \ENDIF
  \IF{SCR or the unbalanced variant}
    \FOR{$n=1,\dots,10$}
      \STATE update $\boldsymbol{\pi}$ by a row scaling, then a column scaling
             \hfill\COMMENT{\eqref{eq:sinkhorn}, approximating \eqref{eq:uot}}
    \ENDFOR
  \ELSE
    \STATE $\boldsymbol{\pi} \leftarrow$ the row or column softmax plan \hfill\COMMENT{\Cref{sec:rules}}
  \ENDIF
\ENDIF
\STATE $m_j \leftarrow \sum_k \pi_{jk}$; \quad $c_k \leftarrow \sum_j \pi_{jk}$
\IF{SCR or the unbalanced variant}
  \STATE $\rho_j \leftarrow Tm_j$; \quad $\sigma_k \leftarrow c_k/b_k$
\ELSE
  \STATE $\rho_j \leftarrow 1$; \quad $\sigma_k \leftarrow 1$
\ENDIF
\STATE $\mathbf{r}_j \leftarrow \rho_j\sum_k (\pi_{jk}/m_j)\,\mathbf{a}_k$
\STATE $\mathbf{V}_j \leftarrow \mathbf{V}_j +
       \mathbf{1}\big(\tanh(\gamma)\,\mathbf{W}\mathbf{r}_j\big)^{\top}$
\STATE $\omega_j \leftarrow \sum_k b_k\sigma_k\pi_{jk}/c_k$
\STATE $\mathbf{s} \leftarrow \sum_j \omega_j\mathbf{v}_j$
\STATE $\mathbf{A} \leftarrow \mathbf{A} +
       \mathbf{1}\big(\tanh(\gamma')\,\mathbf{W}'\mathbf{s}\big)^{\top}$
\end{algorithmic}
\end{algorithm}

\textbf{Marginal residuals of the solver.} Deviation from the prescribed row and column masses is
measured by the relative residuals $\max_j|m_j/a_j-1|$ and $\max_{k:b_k>0}|c_k/b_k-1|$, averaged
over the clips of a fold and then over folds. Across the development models the largest block
mean is below $\RowResidMax$ for the rows and $\ColResidMax$ for the columns, for SCR without the
time prior. For the separate models carrying the prior, the largest row block mean is below
$\PriorRowResidMax$. These bounds apply to the aggregated block means and not to individual folds
or clips.

\section{Backbone Reproduction and Data Protocols}
\label{sec:app-repro}

\subsection{Reproduction and Data Protocols}

A tolerance of $\pm0.5$ UAR, fixed before the runs, defines a successful reproduction.
Against the published scores of \Cref{tab:repro}, single-precision retraining meets the
tolerance for ABVnet on both datasets but not for MMA-DFER, and the released MMA-DFER weights
meet the tolerance. The bfloat16 baselines of the main comparisons fall
\ReproGapLo{} to \ReproGapHi{} UAR below the published models, outside the tolerance, so those
comparisons are internal to this work. The models compared in \Cref{sec:expr} are trained and
evaluated under bfloat16 autocast throughout; the reproduction runs use single precision, as the
original code does. A change of numerical precision and
execution environment, with the configuration otherwise unchanged, shifts the MMA-DFER baseline on DFEW by
\PrecShiftDelta{} UAR in the same direction in all five folds, \PrecShiftCI{}. The fold intervals
do not quantify sensitivity to changes of precision and execution environment, which limits the
interpretation of any single difference in \Cref{tab:confirm}.

\textbf{Input preparation.} MAFW frame files follow string order rather than
numeric order in almost every clip. The released models reproduce their published scores under
string order, which we keep when evaluating the released weights; retraining uses numeric order.
The filterbank is computed from the first audio channel only, following the backbone. The fixed \PadWindow{}
audio window of MMA-DFER starts at the clip onset, and padding occupies \PadShareDfew{} of its
slots on DFEW and \PadShareMafw{} on MAFW; ABVnet uses a shorter window and fewer slots. Face extraction drops frames on DFEW, so
assuming evenly spaced frames misplaces them by more than one slot in \MisplacedClips{} clips.
Frame times are therefore derived from the frame indices in the file names.

\textbf{CREMA-D protocol.} CREMA-D has no official folds, so we define them. Removing the four
clips the dataset documentation lists as faulty leaves 7{,}438. The 91 actors are ordered by sex
and then by ascending actor identifier, the 48 male actors before the 43 female ones. Actor $i$
of that order, counted from zero, is assigned to fold $((i+4)\bmod 5)+1$, so each actor appears
in the test part of exactly one fold. The development protocol holds out whole actors of each
training part, taking them in ascending order of the SHA-256 digest of \texttt{1:} followed by
the actor identifier until a tenth of the clips of that fold is reached. This gives eight actors
and 656 clips in every fold. The assignment is deterministic given the released annotations, with
no random draw at any step.

\begin{table}[!t]
\centering
\caption{Backbone reproduction: mean UAR / WAR over the \SwapFoldCount{} official test folds.
Retraining uses single precision; the main comparisons of \Cref{sec:expr} use bfloat16
autocast.}
\label{tab:repro}
\setlength{\tabcolsep}{4pt}
\fitwidth{%
\begin{tabular}{@{}lcc@{}}
\toprule
Source & DFEW & MAFW \\
\midrule
\multicolumn{3}{@{}l}{\itshape MMA-DFER} \\
Published & 66.61 / 77.15 & 44.19 / 57.90 \\
Released weights\textsuperscript{a} & 66.64 / 77.15 & 44.24 / 57.80 \\
Retrained & 65.42 / 76.76 & 43.60 / 56.84 \\
\midrule
\multicolumn{3}{@{}l}{\itshape ABVnet} \\
Published & 67.52 / 78.28 & 44.46 / 58.41 \\
Released weights & --- & --- \\
Retrained & 67.05 / 77.95 & 44.01 / 57.98 \\
\bottomrule
\end{tabular}}
\par\smallskip
{\footnotesize\raggedright
\textsuperscript{a}Evaluated with 160-pixel inputs and the released frame order.
---: no released weights.\par}
\end{table}

\subsection{Training Configuration and Runtime}
\label{sec:training}

\begin{table}[!t]
\centering
\caption{Backbone inputs, training settings, and the parameters the module adds. The module
variants of \Cref{tab:confirm,tab:dev,tab:axes} share architecture and parameter count, so the
added counts apply to each of them.}
\label{tab:spec}
\setlength{\tabcolsep}{4pt}
\footnotesize
\fitwidth{%
\begin{tabular}{@{}lll@{}}
\toprule
 & MMA-DFER & ABVnet \\
\midrule
\multicolumn{3}{@{}l}{\itshape Inputs, otherwise used as released} \\
Face frames & \multicolumn{2}{l}{$T=16$, sampled over the whole clip} \\
Frame size & $160\times160$ & $224\times224$ \\
Filterbank from onset & 5.12~s & 4.48~s \\
Audio slots & $M=32$ & $M=28$ \\
\midrule
\multicolumn{3}{@{}l}{\itshape Training, following each released recipe} \\
Learning rate & $10^{-4}$ & $4\times10^{-5}$, $3\times10^{-5}$ \\
Weight decay & $10^{-2}$ & $6\times10^{-3}$, $10^{-2}$ \\
Epochs & 25 & 25 \\
Batch size & 8 & 8 \\
\midrule
\multicolumn{3}{@{}l}{\itshape Parameters, over all twelve insertions} \\
Added by the module & 402{,}456 & 5.12M \\
Trainable before & 7.56M & 17.03M \\
Trainable after & 7.96M & 22.16M \\
\bottomrule
\end{tabular}}
\par\smallskip
{\footnotesize\raggedright
The two ABVnet learning rates and decays are for DFEW and MAFW respectively. Both recipes use
AdamW with a cosine schedule to zero and cross-entropy loss, and the gates take no weight
decay. Each insertion carries two LayerNorms, two linear maps and two gates; the trainable
layers of MMA-DFER between the streams are prompts, projections and a joint adaptation layer.
The larger addition in ABVnet follows from the dimension change, each insertion also mapping
both streams between the backbone's 768 dimensions and the module's 128 and back
(\Cref{sec:hosts}), an increase of about 30\%.\par}
\end{table}

\Cref{tab:spec} gives the inputs each backbone takes, the training settings of each
released recipe, and the parameters the module adds. All runs of
\Cref{tab:confirm,tab:dev,tab:axes} use bfloat16 autocast with the module in single precision. Within a table, all
models of a backbone and dataset share the GPU type. On MMA-DFER with DFEW development runs on
one RTX 3090, the median training-epoch time over the \EpochCount{} epochs of five folds was
\EpochBase{}~s for the baseline, \EpochScr{}~s for SCR ($+\EpochScrPct{}\%$) and
\EpochPool{}~s for the pooled variant ($+\EpochPoolPct{}\%$).

\section{Supplementary Results}

\begin{table}[!t]
\centering
\caption{Fold standard deviation of UAR for each configuration of
\Cref{tab:confirm}, over the \SwapFoldCount{} test folds. The comparisons
themselves use paired intervals, which this quantity does not replace.}
\label{tab:foldsd}
\setlength{\tabcolsep}{4pt}
\fitwidth{%
\begin{tabular}{@{}llcccc@{}}
\toprule
Backbone & Dataset & Baseline & SCR & Pooled & \begin{tabular}[b]{@{}c@{}}Column\\softmax\end{tabular} \\
\midrule
MMA-DFER  & DFEW     & 0.73 & 1.30 & 1.06 & 1.55 \\
          & MAFW     & 5.06 & 5.20 & 5.48 & 4.93 \\
          & CREMA-D  & 2.48 & 1.65 & 2.50 & 1.69 \\
\midrule
ABVnet    & DFEW     & 1.82 & 1.62 & 1.45 & 1.43 \\
          & MAFW     & 4.85 & 5.18 & 5.06 & 4.89 \\
\bottomrule
\end{tabular}}
\end{table}

\subsection{Design Axes}

\textbf{Design-axis intervals.} The 90\% confidence intervals of the differences in \Cref{tab:axes} extend \AxesCIHalfLo{} to \AxesCIHalfHi{} UAR on either side of the
mean. Most differences below about one UAR are therefore unresolved and do not rank the
variants.

\begin{table}[!t]
\centering
\caption{Design choices on MMA-DFER with DFEW development data: five-fold mean UAR and paired
$\Delta$UAR, variant minus the pooled variant at the same placement. Differences use unrounded
paired fold values.}
\label{tab:axes}
\setlength{\tabcolsep}{4pt}
\fitwidth{%
\begin{tabular}{@{}lccc@{}}
\toprule
Variant & $\varepsilon$ & UAR & $\Delta$UAR \\
\midrule
\multicolumn{4}{@{}l}{\itshape All twelve blocks} \\
Pooled & --- & 67.27 & --- \\
Column softmax & \epsdef & 67.45 & $+0.17$ (2/5) \\
UOT ($\tau=1$) & \epsdef & 66.91 & $-0.37$ (2/5) \\
SCR & \epsdef & 66.74 & $-0.54$ (2/5) \\
SCR & 0.02 & 66.79 & $-0.49$ (1/5) \\
SCR with time prior & \epsdef & 66.54 & $-0.74$ (2/5) \\
SCR without padding mask & \epsdef & 66.49 & $-0.79$ (2/5) \\
Row softmax & \epsdef & 66.49 & $-0.79$ (1/5) \\
\midrule
\multicolumn{4}{@{}l}{\itshape Blocks 10--12} \\
Pooled & --- & 67.04 & --- \\
Column softmax & 0.02 & 67.36 & $+0.32$ (3/5) \\
SCR & 0.02 & 66.29 & $-0.75^{\dagger}$ (0/5) \\
\bottomrule
\end{tabular}}
\par\smallskip
{\footnotesize\raggedright
Parentheses: folds with higher UAR than the pooled variant.
$\dagger$: the paired 90\% interval excludes zero; no interval is reported for the other
rows.\par}
\end{table}

\textbf{Marginal constraints.} At the default placement, each marginal-constraint variant has an unresolved UAR difference
from the pooled variant (\Cref{tab:axes}); the displayed ordering is one of point estimates.

\textbf{Placement and temperature.} Lowering the temperature to $\varepsilon=0.02$ leaves SCR below
the pooled variant, as at $\varepsilon=\epsdef$. Restricting the module to the last three
blocks, where the interaction is largest (\Cref{tab:routing}), gives SCR a resolved UAR
disadvantage against the pooled variant (\Cref{sec:expr}).

\textbf{Padding mask.} Removing the padding mask reduces the UAR of SCR by \PadMaskDrop{}, an unresolved difference.

\textbf{Backbone design axes.} \Cref{tab:axes} varies the design choices of the module on
MMA-DFER with DFEW development data. Two further axes vary the backbone instead. Replacing the
last-block output by a learned mixture of all twelve blocks reduces mean UAR by \MixDrop{} points in
both the baseline and SCR. The shared reduction is consistent with an effect of the backbone
output choice. Varying the depth of the temporal transformer over one, two and four layers
moves baseline UAR by at most \TempDepthSpread{} points, with no consistent trend. The released
configuration is kept for both models.

\subsection{Per-Class Recall in the Focal Setting}

\begin{table}[!t]
\centering
\caption{ABVnet on DFEW: the module gain is spread unevenly over the classes, with
\PerClassResolved{} resolved per-class difference. Recall differences average the
\SwapFoldCount{} test folds, with \SwapCILevel{} paired-fold intervals against the
retrained baseline. Clips: support over all folds. UAR is the mean of the per-class recalls,
so each column averages to the matching difference in \Cref{tab:comparisons}.}
\label{tab:perclass}
\setlength{\tabcolsep}{4pt}
\footnotesize
\fitwidth{%
\begin{tabular}{@{}lrrcr@{}}
\toprule
 & & \multicolumn{2}{c}{SCR $-$ baseline} & SCR $-$ pooled \\
\cmidrule(lr){3-4}\cmidrule(l){5-5}
Class & Clips & $\Delta$ recall & CI & $\Delta$ recall \\
\midrule
Anger       &  2173 & $+0.51$ & $[-0.28, +1.29]$ & $-0.37$ \\
Disgust     &   145 & $+2.76$ & $[+0.01, +5.51]$ & $+0.69$ \\
Fear        &   902 & $-0.56$ & $[-2.71, +1.59]$ & $-1.88$ \\
Happiness   &  2445 & $+0.61$ & $[-0.10, +1.33]$ & $+0.53$ \\
Neutral     &  2669 & $+1.16$ & $[-1.10, +3.42]$ & $+0.71$ \\
Sadness     &  1894 & $-0.69$ & $[-2.70, +1.33]$ & $-0.05$ \\
Surprise    &  1469 & $+2.32$ & $[-1.31, +5.94]$ & $+1.57$ \\
\bottomrule
\end{tabular}}
\end{table}

Of the \PerClassCount{} classes, \PerClassResolved{} carries a resolved difference from
the retrained baseline: disgust, by \PerClassDisgust{} recall points,
\PerClassDisgustCI{}, on the \PerClassDisgustClips{} clips of the rarest class.
Among \PerClassCount{} comparisons at the \SwapCILevel{} level, fewer than one resolved
difference is expected by chance, so that single class is not protected against selection.
\PerClassDown{} classes move in the opposite direction. Attribution of the gain to
particular classes therefore remains unestablished.

\section{Supplementary Probes}

\subsection{Descriptor, Plan and Update Measurements}

Development clips comprise 938 per fold on DFEW, 733 or 734 on MAFW, 656 on CREMA-D (\Cref{tab:routing}).

The small within-clip spread of the frame descriptors is consistent with those near-separable
costs. In blocks 1 to 9 of MMA-DFER, the relative descriptor deviations, averaged over clips
within folds and then over folds, are at most \FrameSpreadMax{} for the frame descriptors and
\SlotSpreadMin{} or more for the audio-slot descriptors, each relative to the norm of the
corresponding within-clip mean descriptor. This descriptor pattern is consistent with greater cost
variation along the slot axis than along the frame axis, approaching the separable case of
\Cref{cor:separable}.

\begin{figure}[t]
\centering
\includegraphics[width=\columnwidth]{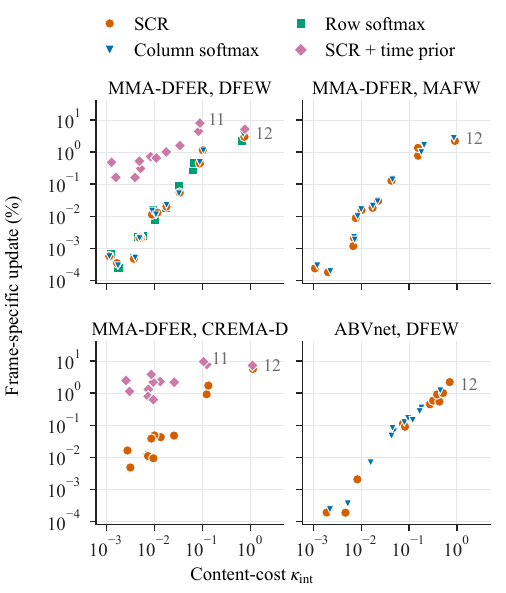}
\caption{Blocks with a larger cost interaction tend to carry a larger frame-specific update;
the association is descriptive and carries no interval. Each point averages clips within a
fold, then folds, for one block of one development model, and the numbers identify blocks. For
ABVnet the update is measured in the latent space of the module.}
\label{fig:kappa}
\end{figure}

The evaluated variants with relaxed marginal constraints also have small measured
frame--slot interactions. Under the column softmax the
block-mean $\kappa_{\mathrm{int}}$ stays at or below \ColKappaMax{} before the final block of
every model trained under the development protocol (\Cref{fig:kappa}). The row softmax
separates concentration from frame-specific routing. In its later blocks the highest-weight
slot receives at least \RowTopSlotMin{} of the row mass, yet its mean frame-specific update
stays below \RowFramePartMax{} of the image tokens before the final block (\Cref{fig:kappa}).
The concentrated slot weights stay similar across frames.

Without the time prior, block means with a larger $\kappa_{\mathrm{int}}$ also tend to carry a
larger mean frame-specific update, across blocks, plan variants, datasets and backbones
(\Cref{fig:kappa}). The two backbones differ in their learned projections, and the
frame-specific update of ABVnet is measured in the space of the module. Nothing here calibrates
$\kappa_{\mathrm{int}}$ against a change in UAR, and the diagnostic is not validated as a
predictor. \Cref{prop:near} gives a first-order expansion of the converged balanced
plan; the proposition does not establish the observed relationship with the gated update
magnitude, and covers neither the implemented SCR plan nor the one-sided softmaxes.

\subsection{Time Prior}

The time prior is an optional ablation of SCR. The cost~\eqref{eq:cost} carries no explicit temporal-distance term: permuting the encoded slot descriptors together with their measure weights permutes the columns of the plan and leaves the routed audio unchanged. The prior penalizes the distance between the frame time $t^{v}_j$ and the slot time $t^{a}_k$,
\begin{equation}
C_{jk} \leftarrow C_{jk} + \lambda\,\big|t^{v}_j - t^{a}_k\big|/\ell ,
\label{eq:timeprior}
\end{equation}
with weight $\lambda=0.2$ and time scale $\ell=1$~s; the cosine distance before this addition is the content cost. The slot time is the center of the slot's 0.16~s window; the frame time is the sampling time of the frame, taken from its index in the source clip rather than assumed uniform (\Cref{sec:app-repro}).

The prior is always applied together with the padding mask. A uniform slot measure would give mass to the padded slots, which lie after the audio of the clip, and the distance penalty would then favor moving that mass onto the last frames.

\textbf{Time prior.} Adding the time prior leaves SCR below the pooled variant on DFEW
(\Cref{tab:axes}), and slightly above the pooled variant on CREMA-D. Against SCR without the prior the paired
difference is \PriorScrDfew{} UAR on DFEW and \PriorScrCrema{} on CREMA-D; against the pooled
variant on CREMA-D the paired difference is \PriorPoolCrema{}. None of these differences is resolved, nor is the
tabulated DFEW difference against pooling.

\subsection{Audio Content and Temporal Routing}
\label{sec:timing}

\begin{table}[t]
\centering
\caption{Audio perturbations on MMA-DFER, DFEW development data: five-fold mean $\Delta$UAR,
perturbed minus unperturbed. Trained models are evaluated without retraining.}
\label{tab:audio}
\setlength{\tabcolsep}{4pt}
\fitwidth{%
\begin{tabular}{@{}lcc@{}}
\toprule
 & \multicolumn{2}{c}{$\Delta$UAR} \\
\cmidrule(l){2-3}
Variant & Shift range & Audio removed \\
\midrule
Baseline & $[+0.07,\,+1.08]$ & $-4.27$ \\
Pooled & $[-0.80,\,+0.18]$ & $-6.92$ \\
SCR & $[-0.06,\,+1.10]$ & $-7.07$ \\
Column softmax & $[-0.01,\,+0.61]$ & $-7.66$ \\
\bottomrule
\end{tabular}}
\par\smallskip
{\footnotesize\raggedright
Shift range: the least and greatest of the eight mean $\Delta$UAR values, one per tested
offset, rounded outward; not a confidence interval. Removal: rounded to nearest.\par}
\end{table}

Training with the time prior makes the routing frame-specific. With the prior the mean
frame-specific update in block 11 reaches \PriorDfewFrameUpdate{} on DFEW and
\PriorCremaFrameUpdate{} on CREMA-D, against at most \FramePartMax{} for SCR without the prior
(\Cref{fig:kappa}). The $\kappa_{\mathrm{int}}$ of the content cost remains comparable to the corresponding
value for SCR without the prior. The frame-specific routing is therefore consistent with the interaction
introduced by the fixed prior.

\textbf{Audio perturbation probes.} The audio is shifted against the video by $\pm80$,
$\pm160$, $\pm320$ and $\pm640$~ms, applied to the filterbank before encoding with the missing
part filled with silence, so the largest shift covers four audio slots. Comparing audio removal across models does not isolate the routing path, since models with and without the module differ in both
capacity and fusion architecture (\Cref{sec:setup}), and measures a contribution at inference
rather than the UAR of a model trained without audio. The shift probes cannot separate unused
timing from tolerance to the offsets tested; filling the shifted part with silence also changes
the audio content.

\section{Solver Iterations and Separability Proofs}
\label{sec:proofs}

\subsection{Sinkhorn Iterations}

\textbf{Sinkhorn iterations.} With dual potentials $\mathbf{f}\in\mathbb{R}^{T}$, $\mathbf{g}\in\mathbb{R}^{M}$ and $\pi_{jk}=a_j b_k \exp\!\big((f_j+g_k-C_{jk})/\varepsilon\big)$, one iteration for~\eqref{eq:uot} is a row step followed by a column step,
\begin{equation}
\begin{aligned}
f_j &\leftarrow -\theta\,\varepsilon \log\textstyle\sum_k b_k\, e^{(g_k - C_{jk})/\varepsilon},\\
g_k &\leftarrow -\theta\,\varepsilon \log\textstyle\sum_j a_j\, e^{(f_j - C_{jk})/\varepsilon},
\end{aligned}
\label{eq:sinkhorn}
\end{equation}
with $\theta=\tau/(\tau+\varepsilon)$, starting from $\mathbf{g}=\mathbf{0}$; SCR uses ten iterations. Each iteration ends with the column step, so with $\theta=1$ the column sums equal $\mathbf{b}$ after every iteration, and the row sums approach $\mathbf{a}$ as the iterations converge. Slots with $b_k=0$ receive no mass, and the arguments below concern the slots with $b_k>0$.

\subsection{Separable Costs}

\textbf{Proof of \Cref{cor:separable}.} The balanced case applies the invariance of the Sinkhorn fixed point to cost terms that do not couple the two indices~\cite{sander2022sinkformers}, specialized to a rectangular cost with prescribed marginals. The balanced argument is given for completeness, followed by the one-sided variants and finite $\tau$. Take $\theta=1$. With $\mathbf{I}=\mathbf{0}$, the Gibbs kernel $K_{jk}=e^{-C_{jk}/\varepsilon} = e^{-\mu/\varepsilon}\,e^{-u_j/\varepsilon}\,e^{-w_k/\varepsilon}$ has rank one. Scaling the rows and columns of a rank-one matrix gives a rank-one matrix $\mathbf{x}\mathbf{y}^{\top}$, and the only such matrix with row sums $\mathbf{a}$ and column sums $\mathbf{b}$ is $\mathbf{a}\mathbf{b}^{\top}$.

For the iterations, the factors $e^{-u_j/\varepsilon}$ and $e^{-w_k/\varepsilon}$ come out of the sums in~\eqref{eq:sinkhorn}. Every row step therefore gives $f_j=\theta u_j+\alpha$, and every column step gives $g_k=\theta w_k+\beta$, for constants $\alpha$ and $\beta$. After one iteration,
\begin{equation*}
\pi_{jk} = e^{(\alpha+\beta-\mu)/\varepsilon}\,a_j b_k\, e^{-(1-\theta)(u_j+w_k)/\varepsilon}.
\end{equation*}
With $\theta=1$ the last factor is one, and the column step makes the constant one, so $\boldsymbol{\pi}=\mathbf{a}\mathbf{b}^{\top}$.

Under the column softmax, $\pi_{jk}= b_k\,\mathrm{softmax}_j(-u_j/\varepsilon)$, and under the row softmax $\pi_{jk}= a_j\,\mathrm{softmax}_k(-w_k/\varepsilon+\log b_k)$; in each case the factor that depends on the other index cancels in the normalization. Substituting the three plans into~\eqref{eq:a2v}, with $\rho_j = T m_j = 1$ for the balanced plan and $\rho_j=1$ for the softmax variants, gives the stated updates.

For $\theta<1$ the factor $e^{-(u_j+w_k)/(\tau+\varepsilon)}$ remains by the identity $(1-\theta)/\varepsilon=1/(\tau+\varepsilon)$. The plan is still the product of a frame factor and a slot factor, so every frame receives the same normalized slot mixture, proportional to $b_k e^{-w_k/(\tau+\varepsilon)}$, scaled by its own mass factor $\rho_j$. This mixture need not equal $\bar{\mathbf{a}}$, and $\rho_j$ need not be one; both approach the balanced values as $\tau$ grows, since the content costs~\eqref{eq:cost} lie in $[0,2]$ and so $|u_j+w_k|\le 4$. \hfill$\blacksquare$

Lowering the temperature therefore cannot make the normalized slot mixture depend on the frame while the cost stays separable.

\subsection{Nearly Separable Costs}

\begin{proposition}[Nearly separable cost]
\label{prop:near}
Let $|I_{jk}|\le\delta$ for every frame and every slot with $b_k>0$. For every $\varepsilon>0$, the converged balanced plan satisfies
\begin{equation}
e^{-4\delta/\varepsilon} \le \frac{\pi_{jk}}{a_j b_k} \le e^{4\delta/\varepsilon}
\label{eq:planbound}
\end{equation}
on that support and, as $\delta/\varepsilon\to0$,
\begin{equation*}
\pi_{jk} = a_j b_k\big(1 - I_{jk}/\varepsilon + O((\delta/\varepsilon)^2)\big).
\end{equation*}
The routed audio then satisfies
\begin{equation*}
\|\mathbf{r}_j-\bar{\mathbf{a}}\| \le \big(e^{4\delta/\varepsilon}-1\big)\textstyle\sum_k b_k\,\|\mathbf{a}_k-\bar{\mathbf{a}}\| .
\end{equation*}
\end{proposition}

The first-order expansion specializes the known large-temperature expansion around the independent coupling~\cite{peyre2026otml}, whose first-order coefficient is this same centered interaction. Entropic transport is also stable more generally~\cite{ghosal2022stability}. \Cref{prop:near} adds the entrywise envelope and its consequence for the routed audio. All of this concerns the converged balanced plan, whereas SCR uses a finite penalty and ten iterations, so it does not directly cover the SCR implementation, which uses a finite penalty and ten iterations;
\Cref{sec:residuals} reports the measured marginal residuals.

For $\delta/\varepsilon\ll1$ the first-order relative deviation $\pi_{jk}/(a_jb_k)-1$ is $-I_{jk}/\varepsilon$, and the weighted root-mean-square size of that term gives the diagnostic~\eqref{eq:kappa}.

\textbf{Proof of \Cref{prop:near}.} Consider the converged balanced plan ($\theta=1$). Absorb $\mu$, $u_j$ and $w_k$ into the scalings, so that $\pi_{jk}=x_j y_k G_{jk}$ with $G_{jk}=e^{-I_{jk}/\varepsilon}\in[e^{-\delta/\varepsilon},e^{\delta/\varepsilon}]$. The row constraint gives $x_j = a_j/\sum_l G_{jl}y_l$, hence $\pi_{jk} = a_j\,G_{jk}y_k/\sum_l G_{jl}y_l$, which lies within a factor $e^{\pm 2\delta/\varepsilon}$ of $a_j\,y_k/\sum_l y_l$; the numerator and the denominator each change by at most a factor $e^{\pm\delta/\varepsilon}$ when every $G_{jk}$ is replaced by one. Summing over $j$ and using the column constraint $\sum_j\pi_{jk}=b_k$ with $\sum_j a_j=1$ shows that $y_k/\sum_l y_l$ lies within a factor $e^{\pm2\delta/\varepsilon}$ of $b_k$, which gives~\eqref{eq:planbound}. For the first-order statement, write $\pi_{jk}=a_jb_k\exp((\phi_j+\psi_k-I_{jk})/\varepsilon)$ and linearize. With $\mathbf{I}$ centered with respect to both measures, the two marginal
constraints force $\phi_j+\psi_k=0$ at first order.

For the update, $\rho_j=Tm_j$ gives $\mathbf{r}_j=T\sum_k\pi_{jk}\mathbf{a}_k$. The rows of the balanced plan sum to $a_j=1/T$, so $\mathbf{r}_j - \bar{\mathbf{a}} = \sum_k (T\pi_{jk}-b_k)(\mathbf{a}_k-\bar{\mathbf{a}})$. By~\eqref{eq:planbound}, $|T\pi_{jk}-b_k|\le(e^{4\delta/\varepsilon}-1)\,b_k$, hence
\begin{equation*}
\|\mathbf{r}_j - \bar{\mathbf{a}}\| \le \big(e^{4\delta/\varepsilon}-1\big)\textstyle\sum_k b_k\,\|\mathbf{a}_k-\bar{\mathbf{a}}\| . \qquad\blacksquare
\end{equation*}

\section*{Data Availability}

DFEW~\cite{jiang2020dfew} and MAFW~\cite{liu2022mafw} are available from their authors for
non-commercial research.
CREMA-D~\cite{cao2014crema} is public under the Open Database License. Our
source code is available at \CodeRepo{}.

\end{document}